\documentclass{article} % For LaTeX2e
\usepackage{iclr2027_conference,times}

\usepackage{amsmath,amsfonts,bm}

\def\eqref#1{equation~\ref{#1}}
\def\1{\bm{1}}

\DeclareMathAlphabet{\mathsfit}{\encodingdefault}{\sfdefault}{m}{sl}
\SetMathAlphabet{\mathsfit}{bold}{\encodingdefault}{\sfdefault}{bx}{n}

\usepackage{hyperref}
\usepackage{url}

\usepackage{booktabs}  

\usepackage{amsmath, amsthm, amssymb}
\newtheorem{proposition}{Proposition}

\usepackage{graphicx}
\usepackage{subcaption}
\usepackage{adjustbox}
\usepackage{multirow}
\usepackage{diagbox}
\usepackage{wrapfig}
\usepackage{enumitem}
\usepackage[table]{xcolor}
\usepackage{xspace}

\usepackage[toc]{appendix}
\usepackage{etoc}
\usepackage{minitoc}

\definecolor{cadmiumgreen}{rgb}{0.0, 0.42, 0.24}
\definecolor{cornellred}{rgb}{0.7, 0.11, 0.11}
\definecolor{citecolor}{HTML}{0071bc}
\definecolor{softpink}{RGB}{166,25,85}

\hypersetup{
  urlcolor = citecolor,
    citecolor = softpink,
  colorlinks = true,
}

\renewcommand{\eqref}[1]{Eq.~\ref{#1}}

\title{What Converges in the Platonic Representation Hypothesis? Structure over Geometry}

\iclrpreprintcopy
\author{
Junwon You$^{1}$, \quad
Mihyun Jang$^{2}$, \quad
Sangwoo Mo$^{2,\dagger}$, \quad
Jae-Hun Jung$^{2,\dagger}$ \\
$^{1}$KAIST \quad
$^{2}$POSTECH \\
{\small
$^{1}$\texttt{jwyou627@gmail.com}
\quad
$^{2}$\texttt{\{jnoodle, sangwoo.mo, jung153\}@postech.ac.kr}
}
}

\begin{document}

\begingroup
\renewcommand*{\thefootnote}{$\dagger$}
\footnotetext[1]{Equal advising. Code: \url{https://github.com/junwon0/what-converges-prh}}
\endgroup
\setcounter{footnote}{0}
\renewcommand{\thefootnote}{\arabic{footnote}}

\maketitle

\begin{abstract}
The Platonic Representation Hypothesis suggests that increasingly capable models converge toward shared representations. 
Recent work narrows this claim to shared local neighborhood relationships, finding that capacity-dependent trends in several global similarity measures largely disappear after calibration.
We challenge this interpretation by showing that prior local-global comparisons confound structural scale (local versus global) with what is compared: relational structure, defined by which samples are related, versus metric geometry, characterized by quantitative relations such as distances, similarities, or correlations.
To disentangle these factors, we construct a controlled $2\times2$ framework that evaluates both relational structure and metric geometry at local and global scales. 
We introduce $H_0$ skeleton overlap as a global counterpart to mutual $k$-nearest neighbors, together with matched distance-aware variants.
Across vision-language models, relational structure exhibits robust representational convergence at both scales after calibration, whereas increasingly stringent distance agreement substantially weakens alignment and progressively flattens the capacity-dependent trend. 
We further extend the analysis beyond ambient Euclidean geometry by evaluating distance agreement under a Riemannian metric approximation and recover the same structure-geometry pattern.
The pattern is also reproduced in video-text representations. 
Together, these results show that relational convergence extends beyond local neighborhoods to global spanning structure, whereas metric geometry exhibits substantially weaker convergence.
\end{abstract}
\section{Introduction}

Whether independently trained neural networks learn similar internal representations has long been a central question in representation learning~\citep{li2015convergent,morcos2018insights,wang2018towards,kornblith2019similarity}.
Existing studies show that independently trained neural networks can share similar representational structures, while also exhibiting substantial variations depending on training conditions and similarity measures~\citep{morcos2018insights,bansal2021revisiting,ding2021grounding}.
As modern models become increasingly capable and heterogeneous, questions of representational convergence have gained renewed importance~\citep{sucholutsky2025getting,huh2024position,tjandrasuwita2025understanding}.
Yet the notion of representational convergence remains underspecified: the key question is not only whether representations become more similar, but what exactly converges.

The Platonic Representation Hypothesis (PRH) proposes that increasingly capable models converge toward a shared representation~\citep{huh2024position}.
Correspondingly, representational alignment tends to increase with model capability across vision and language models~\citep{huh2024position}.
Recent work, however, shows that model scale and layer-wise aggregation can inflate raw similarity, and that aggregation-aware calibration substantially weakens the apparent convergence of several global measures~\citep{groger2026revisiting}.
In contrast, local neighborhood agreement measured by mutual $k$-nearest neighbors (mKNN) remains robust, motivating the view that 
increasingly capable models converge in their local neighborhood relationships~\citep{groger2026revisiting}.

However, we challenge this local neighborhood interpretation because the underlying comparisons differ in structural scale and in whether they compare which samples are related or the numerical relations among them.
mKNN captures local relational structure by comparing which samples are selected as neighbors, without requiring their distances to agree.
In contrast, global similarity measures such as centered kernel alignment (CKA)~\citep{kornblith2019similarity}, CCA-based methods such as SVCCA and PWCCA~\citep{NIPS2017_7188,morcos2018insights}, and representational similarity analysis (RSA)~\citep{kriegeskorte2008representational} depend on numerical relations such as similarities, correlations, or dissimilarities that encode aspects of representation geometry~\citep{williams2021generalized}.
Thus, prior local-global comparisons vary along both dimensions, making it unclear whether the observed difference reflects scale or the distinction between relational structure and metric geometry.
We use \emph{metric geometry} as a broad conceptual category for quantitative relations among samples.

\begin{figure}[!t]
\centering\small
\includegraphics[width=\linewidth]{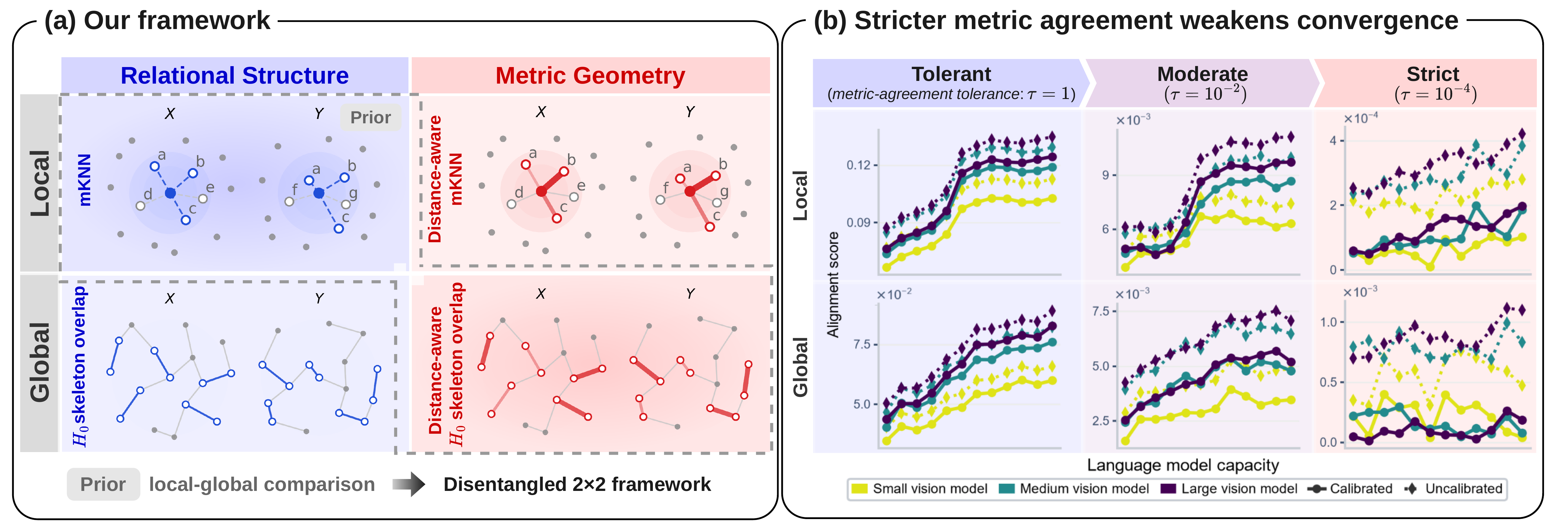} 
\caption{
\textbf{Disentangling structural scale from what is compared.}
(a) Prior local-global comparisons contrast local relational structure with global measures that depend on representation geometry, changing both structural scale and what is compared.
Our controlled $2\times2$ framework disentangles these factors by comparing relational structure and metric geometry independently at local and global scales.
(b) As metric-agreement requirements become more stringent, capacity-dependent convergence weakens at both local and global scales.
Tolerant, moderate, and strict settings correspond to $\tau=1$, $10^{-2}$, and $10^{-4}$ in \eqref{eq:distance_weight}.
}
\label{fig:motiv}
\vspace{-17pt}
\end{figure}

To disentangle these factors, we evaluate representational alignment in relational structure and metric geometry at both local and global scales (Figure~\ref{fig:motiv}\textcolor{red}{a}).
While mKNN captures local relational structure through shared neighbors, we construct a global counterpart using zero-dimensional persistent homology ($H_0$), which captures how disconnected components merge~\citep{edelsbrunner2002topological}.
The corresponding $H_0$ death edges coincide with the edges of a minimum spanning tree (MST)~\citep{skraba2020randomly}, allowing us to compare which edges form the global spanning structure without requiring their distances to agree.
We call this measure \emph{$H_0$ skeleton overlap}.

To probe metric geometry while holding relational structure fixed, we introduce distance-aware variants of both mKNN and $H_0$ skeleton overlap.
These variants additionally require agreement in the distances associated with shared neighbors or spanning edges, with the strictness of this agreement controlled by a parameter $\tau$ in \eqref{eq:distance_weight}.
As $\tau$ decreases, distance agreement becomes increasingly stringent.
We further extend the analysis beyond ambient Euclidean geometry by evaluating distance agreement on the same underlying relations using a Riemannian metric approximation constructed from local covariance information~\citep{singer2008non,berry2016local}.

Across vision-language models, alignment in relational structure remains robust after calibration and generally strengthens with model capacity at both scales, indicating that representational convergence extends beyond local neighborhoods to global spanning structure.
In contrast, as distance-agreement requirements become more stringent, alignment weakens and capacity-dependent trends progressively flatten at both scales, while differences between local and global scales emerge primarily in the most stringent regime.
Extending the analysis beyond Euclidean geometry to a Riemannian metric approximation yields the same structure-geometry pattern.
The pattern is further reproduced in video-text representations, suggesting that it is not specific to a single multimodal setting.

Taken together, our results challenge a local neighborhood interpretation of representational convergence: relational convergence extends to global spanning structure, whereas increasingly stringent metric agreement substantially weakens convergence.

The contributions of this work are as follows:

\begin{itemize}[leftmargin=5mm]

    \item \textbf{Revisiting the local neighborhood interpretation of convergence.}
    We challenge the recent interpretation that calibrated representational convergence is captured by shared local neighborhood relationships.
    We show that the comparisons underlying this view confound structural scale with what is compared: relational structure or metric geometry.

    \item \textbf{Global relational convergence in a controlled 2$\times$2 framework.}
    We introduce a 2$\times$2 framework that compares relational structure and metric geometry at both local and global scales.
    Using $H_0$ skeleton overlap as a global counterpart to mKNN, we show that relational convergence extends beyond local neighborhoods to global spanning structure.

    \item \textbf{Relational structure versus metric geometry.}
    We find that this distinction is more pronounced than the local-global distinction, with scale-dependent differences emerging mainly under stringent agreement.
    Relational structure shows robust convergence at both scales, whereas increasingly stringent metric agreement weakens alignment and progressively flattens the capacity-dependent trend.

    \item \textbf{Robustness beyond Euclidean geometry and across multimodal settings.}
    We extend the controlled distance-aware analysis beyond ambient Euclidean geometry using a Riemannian metric approximation and recover the same structure-geometry pattern.
    We further reproduce this pattern in video-text representations.

\end{itemize}

\section{Related Work}

\subsection{Representational Similarity Measures}

Representational similarity measures quantify agreement between neural representations using different numerical relations among representation vectors.
Methods based on Canonical Correlation Analysis (CCA), including SVCCA and PWCCA~\citep{NIPS2017_7188,morcos2018insights}, compare representation spaces through canonical correlations between linear projections.
Centered Kernel Alignment (CKA)~\citep{kornblith2019similarity} compares kernel-based similarities, while Representational Similarity Analysis (RSA)~\citep{kriegeskorte2008representational} compares pairwise dissimilarities.
Although these measures differ in their invariances and constructions, their scores depend on numerical relations such as correlations, similarities, or dissimilarities.
Relatedly, \citet{williams2021generalized} formulated representation comparison through generalized shape metrics, emphasizing how the choice of comparison measure determines the geometric structure and invariances being compared.

Neighborhood-based approaches capture relational information without requiring exact distances to agree.
Mutual $k$-Nearest Neighbors (mKNN)~\citep{huh2024position}, for example, compares the identities of neighboring samples rather than the corresponding distances.
Other approaches retain ordinal rather than full numerical information.
\citet{soares2026scalable} introduced the Triplet and Quadruplet Similarity Indices (TSI and QSI) to compare relative distance rankings and established a formal connection between TSI and local neighborhood alignment.
More broadly, representational similarity measures preserve and compare different aspects of a representation~\citep{klabunde2025similarity}.
Consequently, an increase in a particular similarity score does not by itself imply convergence in all aspects of representation geometry; interpreting convergence requires identifying which relations a measure preserves and which geometric information it additionally compares.

\subsection{The Platonic Representation Hypothesis and Its Recent Critiques}

The Platonic Representation Hypothesis (PRH)~\citep{huh2024position} proposes that increasingly capable models converge toward a shared statistical representation of the underlying world despite differences in architecture, training data, or modality. Empirical results across multimodal models have supported this view by showing increasing representational alignment with model capability.

However, recent work questions whether raw alignment scores provide direct evidence of representational convergence. \citet{groger2026revisiting} showed that model scale and layer-wise aggregation can inflate similarity scores. Using permutation-based null calibration, they found that scaling trends in CKA, SVCCA, and Procrustes distance largely disappear after calibration, whereas neighborhood-based measures such as mKNN, cycle-kNN, and CKNNA retain clear scaling trends. They further showed that models increasingly agree on local neighbor identities, while the corresponding pairwise distances do not exhibit the same alignment.

This finding establishes an important distinction between relational structure and distance agreement at the local scale, but 
leaves open how to interpret the contrast between robust local neighborhood agreement and weakened global similarity.
In particular, the corresponding measures differ not only in structural scale, but also in what they compare.
Our work addresses this ambiguity by evaluating relational structure and metric geometry at both local and global scales. We pair local mKNN with global $H_0$ skeleton overlap to compare relational structure and construct matched distance-aware variants to assess metric agreement while preserving the underlying relations. 
We further extend this controlled analysis beyond ambient Euclidean geometry using a Riemannian metric approximation.

\subsection{Topological Comparison and Alignment of Representation Spaces}

Persistent homology (PH) has been used to preserve and compare structural information in learned representations, including topology-preserving representation learning~\citep{moor2020topological,kim2024topological} and direct representation comparison through RTD~\citep{barannikov2021representation}.
RTD-Lite further uses minimum spanning trees (MSTs) to efficiently compare connectivity structure across weighted graphs~\citep{tulchinskii2025rtdlite}.
The connection between $H_0$ persistence and MSTs is well established~\citep{kruskal1956shortest,skraba2020randomly}.
Our $H_0$ skeleton overlap discards filtration values and compares only the identities of labeled spanning edges, thereby isolating global relational structure from the distances assigned to those edges.

Topology has also been used as an alignment signal in multimodal representation learning.
Homology Consistency, ToMCLIP, and ToMA use PH-derived structure to align or regularize vision-language representations~\citep{zhang2024homology,you2026topological,you2026topology}.
In contrast, our goal is diagnostic rather than optimization-based: we use $H_0$ skeleton overlap as a global counterpart to local mKNN for comparing relational structure, and construct matched distance-aware variants at both scales to probe metric agreement while preserving the underlying relations.

\section{A Controlled Framework for Representational Convergence}

We develop a controlled framework that separates structural scale from what is compared.
At each scale, we first compare relational structure through the identities of relations selected in each representation, and then probe metric geometry by incorporating distance agreement without changing those relations.
This construction enables matched comparisons at local and global scales while varying whether alignment is evaluated only in relational structure or additionally in the distances assigned to the selected relations.

\subsection{Problem Setup and Comparison Framework}

Let $X=\{x_i\}_{i=1}^n$ and $Y=\{y_i\}_{i=1}^n$ denote two representations of the same $n$ samples, with pairwise distance matrices $D_X$ and $D_Y$, respectively.
We distinguish \emph{structural scale}, separating local relations from global spanning structure.
At each scale, relational structure is represented by the identities of selected relations, whereas metric geometry is probed through the distances assigned to those same relations.
This yields the $2\times2$ framework in Figure~\ref{fig:motiv}\textcolor{red}{a}, allowing us to vary structural scale while holding what is compared fixed, and to introduce distance agreement without changing the underlying relations.

\subsection{Relational Structure Alignment}
\label{sec:relational_alignment}

We first compare relational structure using only the identities of selected relations, without requiring their distances to agree.
We refer to the set of selected relation identities as the structural support.
We use mKNN to capture local relational structure and introduce $H_0$ skeleton overlap as its global counterpart.
We denote all alignment scores by $S(\cdot,\cdot)$, with subscripts specifying the corresponding measure.

\noindent\textbf{Local relational structure: mKNN.}
Following \citet{huh2024position}, let $N_k^X(i)$ and $N_k^Y(i)$ denote the sets of $k$ nearest neighbors of sample $i$ in representations $X$ and $Y$, respectively. We measure local relational alignment by the average overlap between the corresponding neighborhoods:
\begin{equation}
S_{\mathrm{mKNN}}(X,Y)
=
\frac{1}{n}\sum_{i=1}^{n}
\frac{|N_k^X(i)\cap N_k^Y(i)|}{k}.
\label{eq:mknn}
\end{equation}
The score lies in $[0,1]$, with larger values indicating greater agreement in neighbor identities. Importantly, mKNN depends only on which samples are selected as neighbors and does not require the corresponding neighbor distances to agree.

\noindent\textbf{Global relational structure: $H_0$ skeleton overlap.}
To obtain a matched measure of relational structure at the global scale, we use zero-dimensional persistent homology ($H_0$) computed from the Vietoris-Rips filtration~\citep{edelsbrunner2002topological}. 
For each representation, the filtration begins with all samples as separate connected components and progressively adds edges as the distance threshold increases.
Whenever an edge connects two disconnected components, one $H_0$ component disappears; we refer to the corresponding edge as an $H_0$ death edge. Let $E_{H_0}^X$ and $E_{H_0}^Y$ denote the sets of sample pairs corresponding to these death edges in representations $X$ and $Y$, respectively. 

\begin{proposition}
Under a fixed deterministic tie-breaking rule, let $T_X$ and $T_Y$ denote the minimum spanning tree (MST) edge sets of $X$ and $Y$, respectively.
Then the $H_0$ death-edge set of the Vietoris-Rips filtration coincides with the edge set of the MST; that is,
$
E_{H_0}^X = T_X
$
and
$
E_{H_0}^Y = T_Y.
$
\label{prop1}
\end{proposition}

This correspondence follows from Kruskal's construction~\citep{kruskal1956shortest}: an edge produces an $H_0$ death exactly when it connects two previously disconnected components, which is also the criterion for adding an edge to the MST. The connection between persistent homology and minimum spanning structures has also been established more generally~\citep{skraba2020randomly}. 
Formal definitions and a proof of the proposition are provided in Appendix~\ref{app:h0_mst_equivalence}.

Since each MST contains $n-1$ edges, we define $H_0$ skeleton overlap as
\begin{equation}
S_{H_0}(X,Y)
=
\frac{|E_{H_0}^X \cap E_{H_0}^Y|}{n-1}
=
\frac{|T_X \cap T_Y|}{n-1}.
\label{eq:h0_overlap}
\end{equation}
This score measures agreement in the identities of sample pairs that form the global spanning structure. 
We use $H_0$ skeleton overlap as the global counterpart to mKNN, with this local-global distinction formalized in Appendix~\ref{app:h0_global}.
Unlike RTD-Lite, which uses MSTs to quantify multiscale connectivity discrepancies between weighted graphs~\citep{tulchinskii2025rtdlite}, $H_0$ skeleton overlap retains only the identities of the spanning edges, discarding the corresponding edge lengths. 
Together, mKNN and $H_0$ skeleton overlap provide matched measures of relational structure at local and global scales.

\subsection{Distance-Aware Alignment}
\label{sec:distance_aware}

We next probe metric geometry by incorporating distance agreement without changing the relation identities selected at each scale.
This allows us to vary the stringency of distance agreement while holding relational structure fixed.

\noindent\textbf{Distance agreement.}
Given a relation $(i,j)$ shared by the supports of $X$ and $Y$, we evaluate how closely the two representations agree on its distance.
For each representation, pairwise distances are normalized by the 90th percentile of positive distances, $\tilde{D}=D/Q_{0.9}(\{D_{ij}:D_{ij}>0\})$.
We use the 90th percentile as the default reference scale; the resulting patterns are robust to alternative normalization quantiles
(Appendix~\ref{app:distance_normalization}).
We then define the distance-agreement weight
\begin{equation}
w_{ij}^{(\tau)}
=
\exp\left(
-\frac{
\left|
\log \tilde D_X(i,j)
-
\log \tilde D_Y(i,j)
\right|
}{\tau}
\right),
\label{eq:distance_weight}
\end{equation}
where $\tau>0$ controls the stringency of distance agreement.
Smaller $\tau$ requires closer agreement for a shared relation to receive a
large weight, whereas larger $\tau$ is more tolerant of distance mismatch; as
$\tau\to\infty$, all shared relations receive unit weight, recovering the relational structure measure.

\noindent\textbf{Distance-aware mKNN.}
For local alignment, we apply \eqref{eq:distance_weight} only to neighbors shared by the two representations. We define
\begin{equation}
S_{\mathrm{mKNN}}^{(\tau)}(X,Y)
=
\frac{1}{n}
\sum_{i=1}^{n}
\frac{1}{k}
\sum_{j\in N_k^X(i)\cap N_k^Y(i)}
w_{ij}^{(\tau)}.
\label{eq:distance_mknn}
\end{equation}
The neighborhood sets remain identical to those used in mKNN.
Thus, \eqref{eq:distance_mknn} adds distance agreement while preserving the local structural support.

\noindent\textbf{Distance-aware $H_0$ skeleton overlap.}
For global alignment, we apply the same distance-agreement weighting to edges shared by the two $H_0$ skeletons, equivalently, to edges in $T_X\cap T_Y$, and define
\begin{equation}
S_{H_0}^{(\tau)}(X,Y)
=
\frac{1}{n-1}
\sum_{(i,j)\in T_X\cap T_Y}
w_{ij}^{(\tau)}.
\label{eq:distance_h0}
\end{equation}

\noindent\textbf{Controlling distance agreement.}
The parameter $\tau$ controls the strictness of distance agreement: smaller $\tau$ more strongly penalizes a given distance mismatch. Conversely,
\begin{equation}
\lim_{\tau\rightarrow\infty}
S_{\mathrm{mKNN}}^{(\tau)}(X,Y)
=
S_{\mathrm{mKNN}}(X,Y),
\qquad
\lim_{\tau\rightarrow\infty}
S_{H_0}^{(\tau)}(X,Y)
=
S_{H_0}(X,Y).
\label{eq:support_limit}
\end{equation}
Since $w_{ij}^{(\tau)}\to 1$ as $\tau\to\infty$, both distance-aware measures recover their corresponding relational structure measures.
To examine whether the analysis depends specifically on distance-based comparison, we construct analogous similarity-aware variants using cosine similarity while keeping the same structural supports fixed (Appendix~\ref{app:similarity_values}).

\subsection{Aggregation-Aware Permutation Calibration}
\label{sec:per_cal}

We evaluate all alignment measures using the aggregation-aware permutation calibration of \citet{groger2026revisiting}. For each model pair, sample correspondences are randomly permuted using the same permutation across all layers. We take the maximum scores across layer pairs, ensuring that the null distribution accounts for layer selection and aggregation.

Let $S(X_\ell,Y_{\ell'})$ denote the alignment score between layers $\ell$ and $\ell'$. 
The observed aggregate alignment and the corresponding aggregate scores under $K$ random permutations $\{\pi_r\}_{r=1}^K$ are
\begin{equation}
T_{\mathrm{obs}}
=
\max_{\ell,\ell'} S(X_\ell,Y_{\ell'}), \qquad
T^{(r)}
=
\max_{\ell,\ell'} S(X_\ell,\pi_r(Y_{\ell'})).
\label{eq:observed_aggregate}
\end{equation}
We define $c_\alpha$ as the $(1-\alpha)$ quantile of
$\{T_{\mathrm{obs}},T^{(1)},\ldots,T^{(K)}\}$. Since all measures in our framework are bounded above by $1$, the calibrated score is
\begin{equation}
T_{\mathrm{cal}}
=
\max\left\{
\frac{T_{\mathrm{obs}}-c_\alpha}
{1-c_\alpha},
0
\right\}.
\label{eq:calibrated_score}
\end{equation}
Thus, scores that do not exceed the calibration threshold are mapped to zero, while the maximum possible alignment remains one. We apply the same aggregation and calibration procedure to all alignment measures in our framework. Additional details on the number of permutations, permutation significance tests, and multiple-testing correction are provided in Appendix~\ref{app:statistics}. 

\subsection{Beyond Euclidean Geometry}
\label{sec:method_riemannian}

Our primary distance-aware analysis evaluates metric agreement using ambient Euclidean distances.
To extend the analysis beyond ambient Euclidean geometry, we construct a Riemannian metric approximation from local covariance information, motivated by local covariance-based geometric constructions~\citep{singer2008non,berry2016local}.
We keep the relation identities used in the Euclidean analysis fixed and change only the distances used to evaluate agreement on those relations.
This provides a controlled extension in which relational structure remains unchanged while the geometry used to evaluate metric agreement varies.
We treat this construction as one alternative geometric model rather than as a uniquely correct intrinsic geometry of the representation space.
Construction details are provided in Appendix~\ref{app:riemannian_method}.

\section{Experiments}

Our experiments are designed to answer the following key research questions:
\vspace{-5pt}
\begin{itemize}[leftmargin=3mm]

    \item \textbf{Relational structure:}
    Does relational convergence extend from local neighborhoods to global spanning structure?
    (Section~\ref{sec:ex_support})
    \vspace{-3pt}
    
    \item \textbf{Metric geometry:}
    How does convergence change with stricter distance agreement?
    (Section~\ref{sec:ex_value})
    \vspace{-3pt}

    \item \textbf{Beyond Euclidean geometry:}
    Does the structure-geometry pattern persist under a Riemannian metric approximation? (Section~\ref{sec:ex_riemannian})
    \vspace{-3pt}

    \item \textbf{Multimodal generalization:}
    Does the pattern extend to video-text representations?
    (Section~\ref{sec:video_text})
    \vspace{-3pt}

\end{itemize}

\vspace{-8pt}

\subsection{Experimental Setup}
\label{sec:setup}

\noindent\textbf{Vision-language setting.}
We use the 1,024 paired image-text samples from the WIT subset of the PRH benchmark~\citep{huh2024position,groger2026revisiting}.
Our main evaluation compares 12 language models from the BLOOMZ, OpenLLaMA, and LLaMA families with 17 vision transformers spanning ImageNet-21K supervised models, MAE, DINOv2, and CLIP variants, yielding 204 model pairs.
Full model, layer, and representation-extraction details are provided in Appendix~\ref{app:vl_details}.

\noindent\textbf{Alignment and calibration.}
We use $k=10$ for mKNN and evaluate the distance-aware measures over
$\tau\in \{1,0.1,0.01,0.001,0.0001\}$.
Following Section~\ref{sec:per_cal}, model-pair alignment is obtained by taking the maximum over all layer pairs and calibrated using 500 correspondence permutations at $\alpha=0.05$.
Unless otherwise stated, all reported alignment values are aggregation-aware calibrated scores.
We control the false discovery rate across model pairs using the Benjamini-Hochberg procedure~\citep{benjamini1995controlling}, with full statistical testing details provided in Appendix~\ref{app:multiple_testing}.

\subsection{Relational Structure Converges at Both Local and Global Scales}
\label{sec:ex_support}

\begin{figure}[!t]
\centering\small
\includegraphics[width=0.99\linewidth]{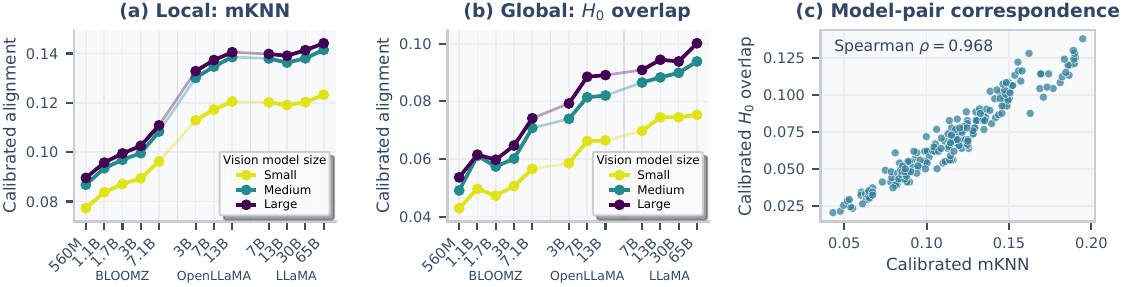} 
% \vspace{-6pt}
\caption{
\textbf{The relational structure of representations also converges globally.}
(a) Following prior work, aggregation-aware calibrated mKNN reproduces the capacity-dependent convergence.
(b) Our $H_0$ skeleton overlap shows that the relational structure of representations also converges at the global scale, extending to global spanning structure.
Small, medium, and large denote relative model sizes within each vision-model family, averaged across INet21K, MAE, DINOv2, CLIP, and CLIP fine-tuned on ImageNet-12K.
(c) Across all 204 vision-language model pairs, calibrated mKNN and $H_0$ skeleton overlap are strongly associated (Spearman $\rho=0.968$).
}
\label{fig:support_convergence}
\vspace{-8pt}
\end{figure}

We first ask whether the convergence in local neighborhood structure reported by prior work extends to global spanning structure after aggregation-aware calibration.
All 204 model pairs remain significant for both mKNN and $H_0$ skeleton overlap after Benjamini-Hochberg correction (Appendix~\ref{app:value_sensitive_sweep}).
Full family-wise plots are provided in Appendix~\ref{app:support}.

Figures~\ref{fig:support_convergence}\textcolor{red}{a} and~\ref{fig:support_convergence}\textcolor{red}{b} show that the two measures of relational structure exhibit similar scaling behavior.
Within language-model families, calibrated alignment generally increases with model capacity, and larger vision models tend to exhibit stronger alignment.
Thus, the capacity-dependent pattern observed for local neighborhood relations also extends to global spanning structure.

This correspondence also holds at the level of individual model pairs.
Across all 204 vision-language pairs, calibrated mKNN and $H_0$ skeleton overlap are strongly correlated (Spearman $\rho=0.968$; Figure~\ref{fig:support_convergence}\textcolor{red}{c}).
Model pairs with stronger local relational alignment therefore also tend to exhibit stronger global relational alignment.
Together, these results indicate that convergence in relational structure is not restricted to local neighborhoods; it extends to global spanning structure even after aggregation-aware calibration.

\subsection{Stricter Distance Agreement Weakens Capacity-Dependent Convergence}
\label{sec:ex_value}

\begin{figure}[!t]
\centering\small
\includegraphics[width=0.99\linewidth]{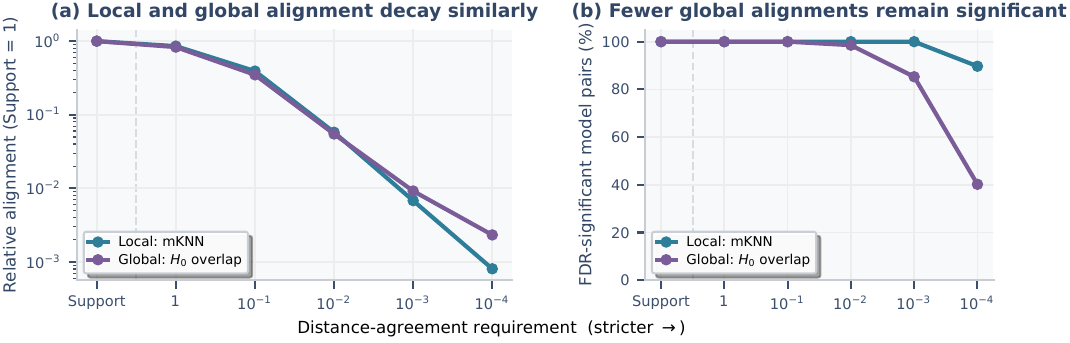}
% \vspace{-6pt}
\caption{
\textbf{The metric geometry of representations shows weaker convergence at both local and global scales.}
(a) Mean aggregation-aware calibrated alignment across the 204 model pairs, normalized by the corresponding relational structure baseline, which is set to 1.
Here, “Support'' denotes this baseline without distance weighting.
As distance agreement becomes more stringent, alignment weakens at similar rates for both local and global scales.
(b) Percentage of the 204 model pairs that remain significant after Benjamini-Hochberg correction at each level of distance agreement.
Significance remains similar across scales under weak-to-moderate agreement, while fewer global alignments remain significant in the stringent regime.
}
\label{fig:value_sensitivity}
\vspace{-9pt}
\end{figure}

Having established robust convergence in relational structure at both scales, we next ask how this pattern changes when increasingly stringent distance agreement is required.
Starting from the relational structure measures in Section~\ref{sec:relational_alignment}, we progressively impose distance agreement through the parameter $\tau$ defined in Section~\ref{sec:distance_aware}.
Because decreasing $\tau$ assigns smaller weights to a fixed distance mismatch, absolute alignment is expected to decrease. The more substantive question is whether the capacity-dependent signature of convergence is preserved as distance agreement becomes more stringent, and whether this behavior differs between local and global scales.

Figure~\ref{fig:value_sensitivity}\textcolor{red}{a} shows that local and global alignment weaken at remarkably similar rates.
At $\tau=1$, the local and global measures retain 85.6\% and 83.0\% of their relational structure baselines, respectively.
These values decrease to 39.2\% and 34.9\% at $\tau=10^{-1}$, and to 5.8\% and 5.5\% at $\tau=10^{-2}$.
Thus, over the weak-to-moderate regime, requiring closer distance agreement produces nearly parallel reductions in alignment at the two structural scales.

More importantly, increasingly stringent distance agreement weakens the capacity-dependent signature observed in relational structure.
Across the distance-aware sweep, this pattern becomes less pronounced as the agreement requirement becomes more stringent (Appendix~\ref{app:value_sensitive_sweep}).
Exact family-level permutation tests of within-family Spearman associations confirm this weakening for both language and vision capacity at both local and global scales (all Benjamini-Hochberg adjusted $q<0.05$; Appendix~\ref{app:capacity_trend}).
This indicates that convergence with increasing model capacity is less robust when metric agreement is additionally required.

A scale-dependent difference emerges under the most stringent distance requirements. As shown in Figure~\ref{fig:value_sensitivity}\textcolor{red}{b}, the prevalence of significant local and global alignment remains nearly identical under weak-to-moderate agreement, but diverges as the requirement becomes stringent. All 204 pairs remain significant for local alignment through $\tau=10^{-3}$, whereas significant global alignment decreases to 98.5\% at $\tau=10^{-2}$, 85.3\% at $\tau=10^{-3}$, and 40.2\% at $\tau=10^{-4}$. At the strictest setting, 89.7\% of local alignments remain significant, compared with 40.2\% of global alignments. 

Together, these results show that increasingly stringent distance agreement weakens convergence at both local and global scales.
The dominant contrast is therefore between relational structure and metric geometry, while structural scale becomes more consequential mainly in the stringent regime, where significant global alignment is less prevalent than local alignment.
The same qualitative weakening is also observed when agreement is defined using cosine similarity rather than distance (Appendix~\ref{app:similarity_values}).
Full numerical results and family-wise analyses are provided in Appendix~\ref{app:value_sensitive}.

\vspace{-2pt}

\subsection{Extending Beyond the Ambient Euclidean Metric}
\label{sec:ex_riemannian}

We next extend the distance-aware analysis beyond the ambient Euclidean metric.
Following Section~\ref{sec:method_riemannian}, we keep the local and global relation identities fixed and evaluate distance agreement using a Riemannian metric approximation constructed from local covariance information.
This preserves the underlying relational structure while changing the geometry used to evaluate metric agreement.

\begin{figure}[!t]
\centering\small
\includegraphics[width=0.99\linewidth]{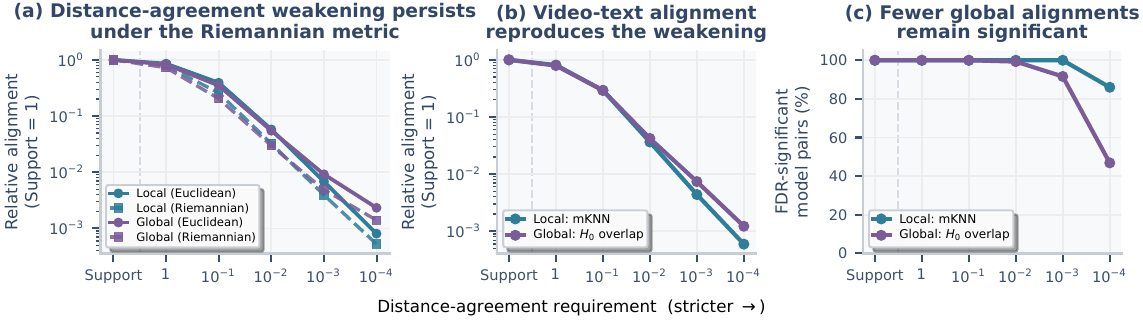}
% \vspace{-6pt}
\caption{
\textbf{The structure-geometry pattern extends beyond Euclidean geometry and to video-text representations.}
(a) Relative calibrated alignment under the ambient Euclidean metric and a Riemannian metric approximation.
The underlying relations are held fixed while only the geometry used to evaluate distance agreement changes.
(b) The same weakening under increasingly stringent distance agreement is reproduced in video-text representations at both local and global scales.
(c) Percentage of video-text model pairs that remain significant after Benjamini-Hochberg correction at each level of distance agreement.
}
\label{fig:robustness_replication}
\vspace{-8pt}
\end{figure}

As shown in Figure~\ref{fig:robustness_replication}\textcolor{red}{a}, local and global alignment again decrease at similar rates as $\tau$ becomes smaller.
At $\tau=10^{-2}$, only $3.2\%$ and $3.0\%$ of the corresponding relational structure baselines remain for the local and global measures, respectively.
More importantly, the full capacity-dependent sweeps show that the increase in alignment with model capacity becomes progressively less pronounced as Riemannian distance agreement becomes more stringent (Appendix~\ref{app:riemannian_results}).

Thus, the weakening of capacity-dependent convergence under stringent metric agreement is not specific to ambient Euclidean geometry.
The same structure-geometry pattern emerges under the Riemannian metric approximation.

\subsection{Extension to Video-Text Representations}
\label{sec:video_text}

Finally, we test whether the observed pattern extends to video-text representations.
Experimental details are provided in Appendix~\ref{app:video_setup}.
Convergence in relational structure is again robust at both scales: all 143 video-text model pairs exhibit significant calibrated alignment for both mKNN and $H_0$ skeleton overlap.
The corresponding family-wise results show similar capacity-dependent patterns across VideoMAE, DINOv2, and CLIP (Appendix~\ref{app:video_text}).

Figure~\ref{fig:robustness_replication}\textcolor{red}{b} shows that increasingly stringent distance agreement again weakens both local and global alignment.
The full capacity-dependent sweeps further show that the scaling trend progressively flattens as distance agreement becomes more stringent (Appendix~\ref{app:video_text}).

A scale-dependent difference again emerges under stringent distance agreement.
At $\tau=10^{-4}$, $86.0\%$ of local alignments remain significant after Benjamini-Hochberg correction, compared with $46.9\%$ of global alignments (Figure~\ref{fig:robustness_replication}\textcolor{red}{c}).
Thus, the same structure-geometry pattern extends to the video-text setting.
\section{Conclusion}

Our findings challenge a local neighborhood interpretation of representational convergence.
When structural scale is separated from what is compared, relational structure converges at both local and global scales, whereas increasingly stringent metric agreement substantially weakens convergence.
This suggests that the key question is not simply whether representations converge locally or globally, but what aspects of their structure are shared across models.
These findings also motivate representation alignment methods that consider both local neighborhoods and global spanning structure, rather than focusing on local structure alone.

Our analysis is limited to $H_0$ spanning structure, a fixed-relation formulation of metric agreement, and a single Riemannian metric approximation.
The $H_0$ skeleton captures only one form of global relational structure, and other global structures may exhibit different convergence patterns.
Future work should examine richer topological structures, alternative formulations of metric agreement, and other approximations of intrinsic representation geometry.

\bibliography{iclr2027_conference}
\bibliographystyle{iclr2027_conference}

\clearpage
\hypersetup{linkcolor=black}
\etocdepthtag.toc{mtappendix}
\etocsettagdepth{mtchapter}{none}
\etocsettagdepth{mtappendix}{subsection}
\tableofcontents
\hypersetup{linkcolor=red}

\appendix
\appendix

\section{Experimental Details}
\label{app:experimental_details}

\subsection{Vision-Language Models and Representations}
\label{app:vl_details}

Our vision-language experiments follow the experimental protocol of \citet{huh2024position}, as adopted by \citet{groger2026revisiting}. We use 1,024 paired image-text samples from the WIT (Wikipedia-based Image Text) dataset~\citep{srinivasan2021wit}. We evaluate language and vision models at multiple scales to examine representational alignment across model capacity.

For language models, we consider 12 models spanning three model families: BLOOMZ~\citep{muennighoff2023crosslingual}, OpenLLaMA~\citep{openlm2023openllama}, and LLaMA~\citep{touvron2023llama}. For vision models, we consider 17 Vision Transformers spanning ImageNet-21K supervised models~\citep{dosovitskiy2021an, steiner2022how}, MAE~\citep{he2022masked}, DINOv2~\citep{oquab2024dinov}, and CLIP variants~\citep{radford2021learning, cherti2023reproducible}. The resulting combination yields 204 vision-language model pairs.

For each language model, we extract representations from all available hidden states and mean-pool over non-padding tokens. For each vision model, we extract representations from every transformer block using the CLS token. Thus, each model is represented by a sequence of layer-wise feature vectors over the same 1,024 samples.

Before computing alignment measures, we apply the same feature preprocessing procedure to all representations. 
Specifically, for each model representation, we compute the 95th percentile of the absolute feature values for each sample and average these sample-wise quantiles to obtain a single clipping threshold. Feature values are then clipped symmetrically to this threshold, and the resulting layer-wise representations are L2-normalized before constructing neighborhood and spanning supports.

Because the layer at which cross-model alignment occurs is not known a priori, we compute the alignment score for every pair of layers between a vision model and a language model and take the maximum across layer pairs as the model-pair alignment. This aggregation procedure is applied consistently across all alignment measures.

We evaluate alignment in relational structure at both local and global scales.
For local relational structure, we use mKNN with $k=10$, while for global relational structure, we use $H_0$ skeleton overlap.
We then evaluate distance-aware variants of both measures by imposing increasingly stringent distance agreement through
$\tau \in \{1,0.1,0.01,0.001,0.0001\}$.

For the distance-aware analysis, the structural supports are first constructed from the normalized representations and then held fixed while distance agreement is evaluated on the corresponding relations.
Our primary analysis uses the ambient Euclidean metric.
We additionally repeat the analysis using a locally estimated Riemannian metric while keeping the underlying structural supports fixed, testing whether the observed structure-geometry pattern depends on the choice of metric.

\subsection{Video-Text Experimental Setup}
\label{app:video_setup}

We additionally test whether the observed alignment patterns extend beyond vision-language representations by repeating the same analysis in a video-text setting. We use 1,024 video samples from the test split of PVD~\citep{bolya2025perception,cho2025perceptionlm}.

For video-native representations, we use VideoMAE~\citep{tong2022videomae}. We evaluate a scale series consisting of the Base and Large pretrained checkpoints and the Huge checkpoint fine-tuned on Kinetics. This provides video representations spanning multiple model scales and fine-tuning conditions.

As a frame-level baseline, we apply image-based vision models to the middle frame of each video. Specifically, we use DINOv2 (small, base, large, and giant) and CLIP (base, large, huge, and giant) to extract representations from the selected middle frame. These models provide a comparison between video-native representations and representations obtained from a single image frame.

VideoMAE representations are extracted from 16-frame clips using the CLS token at each hidden layer. For sufficiently long videos, we use up to four 16-frame clips obtained from uniformly sampled frames and average the resulting layer-wise representations across clips. DINOv2 and CLIP operate on a single middle frame, with CLS representations extracted from every transformer block.

The language models include the same 12 models used in the vision-language experiments together with Gemma-2-9B-IT~\citep{team2024gemma}, spanning the BLOOMZ, OpenLLaMA, LLaMA, and Gemma families. Language representations are extracted using the same procedure described in Appendix~\ref{app:vl_details}. The resulting combinations yield 143
video-text model pairs.

We evaluate video-text alignment in relational structure at both local and global scales using mKNN and $H_0$ skeleton overlap, respectively.
We further evaluate distance-aware variants of both measures under increasingly stringent distance agreement, with
$\tau \in \{1,0.1,0.01,0.001,0.0001\}$.
As in the vision-language experiments, alignment is obtained by taking the maximum across all layer pairs.
The same preprocessing, structural-support construction, distance-aware evaluation, and permutation-based calibration procedures are applied to the video-text experiments.

\section{Calibration and Statistical Testing}
\label{app:statistics}

\subsection{Aggregation-Aware Permutation Calibration}
\label{app:calibration_details}

We use permutation-based null calibration~\citep{groger2026revisiting} to account for alignment that may arise by chance under finite-sample and high-dimensional settings, as well as inflation introduced by layer-wise aggregation.
For each pair of models, we first compute the alignment score for every pair of layers and construct the corresponding layer-pair alignment matrix.
The observed aggregate alignment is then obtained using the same aggregation operator employed throughout our experiments, namely the maximum over all layer pairs.

Formally, let $X_\ell$ and $Y_{\ell'}$ denote the representations of layers $\ell$ and $\ell'$ from two models.
For an alignment measure $S$, we construct the observed alignment matrix
\begin{equation}
\mathbf{S}_{\mathrm{obs}}
=
\left[
S(X_\ell,Y_{\ell'})
\right]_{\ell,\ell'}.
\end{equation}
The observed aggregate statistic is
\begin{equation}
T_{\mathrm{obs}}
=
\max_{\ell,\ell'}
S(X_\ell,Y_{\ell'}).
\end{equation}

To construct an empirical null distribution, we randomly permute the sample correspondences between the two models.
For each permutation $\pi_r$, we apply the same permutation to every layer of the second model and recompute the complete layer-pair alignment matrix:
\begin{equation}
\mathbf{S}^{(r)}
=
\left[
S(X_\ell,\pi_r(Y_{\ell'}))
\right]_{\ell,\ell'}.
\end{equation}
We then apply the same aggregation operator to obtain the permuted aggregate statistic
\begin{equation}
T^{(r)}
=
\max_{\ell,\ell'}
S(X_\ell,\pi_r(Y_{\ell'})),
\qquad r=1,\ldots,K.
\end{equation}

Applying a common permutation across all layers preserves sample correspondence across layers within each model while disrupting cross-model correspondence.
Moreover, taking the maximum after permutation reproduces the same layer-selection and aggregation procedure used for the observed statistic.
The resulting null distribution therefore accounts for inflation induced by selecting the maximum across multiple layer pairs.

We use $K=500$ random permutations for each model pair and set the significance level to $\alpha=0.05$ throughout the experiments.
Let
$
\mathcal{T} = \left\{ T_{\mathrm{obs}}, T^{(1)},\ldots,T^{(K)}
\right\}
$
denote the combined set of observed and permuted aggregate statistics.
We sort these values in ascending order as
$
T_{(1)} \leq \cdots \leq T_{(K+1)}.
$
The permutation critical value is defined as the empirical right-tail $(1-\alpha)$ quantile:
\begin{equation}
c_\alpha
=
T_{\left(
\left\lceil
(1-\alpha)(K+1)
\right\rceil
\right)}.
\label{eq:app_critical_value}
\end{equation}
This value provides the reference level for calibrating observed alignment relative to the permutation null.

For alignment measures bounded above by one, we transform the observed aggregate statistic into a calibrated score as
\begin{equation}
T_{\mathrm{cal}}
=
\max\left\{
\frac{T_{\mathrm{obs}}-c_\alpha}
{1-c_\alpha},
0
\right\}.
\label{eq:app_calibrated_score}
\end{equation}
This transformation maps scores at or below the calibration threshold to zero while preserving an upper bound of one.

The same permutation scheme, layer-wise aggregation, critical-value construction, and calibration transformation are applied to all alignment measures considered in our experiments.
This provides a common calibration procedure across relational structure and distance-aware analyses.

\subsection{Permutation Significance Tests and Multiple-Testing Correction}
\label{app:multiple_testing}

In addition to calibrated alignment scores, we perform permutation-based significance tests to determine whether the observed aggregate alignment exceeds that expected under disrupted cross-model sample correspondence.
Using the same permutation aggregate statistics, we compute the one-sided empirical permutation $p$-value
\begin{equation}
p
=
\frac{
1+
\sum_{r=1}^{K}
\mathbb{I}
\left[
T^{(r)} \geq T_{\mathrm{obs}}
\right]
}
{K+1},
\label{eq:app_permutation_pvalue}
\end{equation}
where $\mathbb{I}[\cdot]$ is the indicator function.
The add-one correction yields a valid finite-sample permutation $p$-value and prevents the reported value from being exactly zero.

At the nominal level, a model pair satisfies $p\leq0.05$ when its observed aggregate alignment is unusually large relative to the permutation distribution.
Because multiple model pairs are tested, however, our reported significance decisions are based on false discovery rate control rather than the uncorrected threshold.

We use the Benjamini-Hochberg procedure~\citep{benjamini1995controlling}.
Suppose that $m$ model-pair comparisons yield permutation $p$-values $p_1,\ldots,p_m$, sorted as
\[
p_{(1)} \leq p_{(2)} \leq \cdots \leq p_{(m)}.
\]
For a target FDR level $q$, we identify the largest index $j$ satisfying
\begin{equation}
p_{(j)}
\leq
\frac{j}{m}q.
\label{eq:app_bh}
\end{equation}
All hypotheses corresponding to $p_{(1)},\ldots,p_{(j)}$ are then declared significant.
We set $q=0.05$ and apply the Benjamini-Hochberg correction separately across model-pair tests for each alignment measure and each $\tau$ setting.
The resulting FDR-controlled decisions are used to report statistically significant alignment, while calibrated scores quantify alignment above the permutation-based calibration threshold.

\section{\texorpdfstring{$H_0$}{H0} Skeleton and Minimum Spanning Trees}
\label{app:h0_mst}

\subsection{Formal Definition of the \texorpdfstring{$H_0$}{H0} Skeleton}
\label{app:h0_definition}

Let
\[
X=\{x_1,\ldots,x_n\}
\]
be a representation of $n$ labeled samples, and let $D_X$ denote its
pairwise distance matrix.

Consider the complete weighted graph
\[
G_X=(V,E,w_X),
\qquad
V=\{1,\ldots,n\},
\qquad
w_X(i,j)=D_X(i,j),
\]
where each vertex index identifies the same sample across representations.
Let
\[
e_1 \prec_X e_2 \prec_X \cdots \prec_X e_m,
\qquad
m=\binom{n}{2},
\]
denote a fixed total ordering of the edges that is nondecreasing in
$w_X$. Ties are resolved deterministically as described in
Appendix~\ref{app:h0_ties}.

For $r=0,\ldots,m$, let
\[
G_X^{(r)}
=
\bigl(V,\{e_1,\ldots,e_r\}\bigr),
\]
with $G_X^{(0)}$ containing only the $n$ isolated vertices.
This ordered graph filtration is a tie-refined $1$-skeleton filtration
of the Vietoris-Rips filtration.
An edge $e_r=(i,j)$ produces an $H_0$ death event precisely when its
endpoints belong to distinct connected components of $G_X^{(r-1)}$.
We therefore define the labeled $H_0$ death-edge set as
\[
E_{H_0}^{X}
=
\left\{
e_r=(i,j)\in E :
i \text{ and } j
\text{ belong to distinct connected components of }
G_X^{(r-1)}
\right\}.
\]

The corresponding $H_0$ skeleton is the labeled graph
\[
\mathcal{H}_0(X)
=
\bigl(V,E_{H_0}^{X}\bigr).
\]
Each edge in $E_{H_0}^{X}$ records the sample pair responsible for
merging two previously disconnected components.
Thus, the $H_0$ skeleton retains the identities of the relations that
form the spanning connectivity structure while discarding their
filtration values.
In Section~\ref{sec:relational_alignment}, we compare these labeled edge
identities across representations to measure global relational alignment.

\subsection{\texorpdfstring{$H_0$}{H0} Death Edges and Minimum Spanning Trees}
\label{app:h0_mst_equivalence}

We now formalize the correspondence between $H_0$ death edges and
minimum spanning trees.
Let $\prec_X$ be any total ordering of the edges that is consistent with
their weights, with ties resolved by a fixed deterministic rule.
Let $T_X$ denote the edge set returned by Kruskal's
algorithm~\citep{kruskal1956shortest} under this ordering.

\paragraph{Proposition 1.}
Under the same total ordering $\prec_X$, the labeled $H_0$ death-edge set
coincides with the minimum spanning tree edge set:
\[
E_{H_0}^{X}=T_X.
\]

\paragraph{Proof of Proposition 1.}
Let $F_X^{(r)}$ denote the forest produced by Kruskal's algorithm after
processing the first $r$ edges.
We first observe that $F_X^{(r)}$ and $G_X^{(r)}$ have the same connected
components for every $r$.
Initially, both contain $n$ isolated vertices.
When an edge $e_r=(i,j)$ is processed, either its endpoints are already
connected, in which case neither graph changes its connected components,
or they lie in distinct components, in which case $G_X^{(r)}$ merges
those components and Kruskal's algorithm accepts $e_r$, producing the
same merge.

Therefore,
\[
e_r\in E_{H_0}^{X}
\quad\Longleftrightarrow\quad
i \text{ and } j \text{ are disconnected in } G_X^{(r-1)}
\quad\Longleftrightarrow\quad
e_r\in T_X.
\]
Hence $E_{H_0}^{X}=T_X$.
Since the complete weighted graph is connected, the resulting tree
contains exactly $n-1$ edges:
\[
|E_{H_0}^{X}|=|T_X|=n-1.
\]
The same argument applies to $Y$.
\hfill$\square$

\subsection{Tie Handling and Interpretation}
\label{app:h0_ties}

When tied edge weights admit multiple valid minimum spanning trees
(MSTs), our implementation selects a deterministic representative using
a dense implementation of Prim's algorithm~\citep{prim1957shortest}.
The algorithm is initialized at the first sample; among candidates with
the same current minimum weight, the lowest-index sample is selected,
and an existing parent is updated only when a strictly smaller edge
weight is encountered.
Since sample indices are fixed across representations, this procedure
yields a deterministic labeled MST for each representation.

This implementation is consistent with the correspondence established in
Appendix~\ref{app:h0_mst_equivalence}.
The deterministic MST selected by Prim's algorithm can be represented as
the MST obtained by Kruskal's algorithm under an appropriate deterministic
refinement of tied edge weights.
Under that refinement, the selected MST therefore coincides with the
corresponding labeled $H_0$ death-edge set.
When all pairwise weights are distinct, the MST is unique and no tie
refinement is required.

Although individual MST edges may connect nearby samples, the edge set is
globally constrained by the requirement of spanning all samples without
cycles.
The resulting $n-1$ labeled edges therefore encode a connectivity
structure over the entire representation.
We interpret this edge set as global spanning relational structure,
without implying that individual edges are geometrically long-range or
that the $H_0$ skeleton captures the full global geometry.

\subsection{Global Nature of the \texorpdfstring{$H_0$}{H0} Skeleton}
\label{app:h0_global}

The distinction between local and global structure in our framework can be
understood through connectivity over the point cloud. A $k$-nearest-neighbor
graph is constructed from sample-centered neighborhoods, so its relations are
determined from the $k$ nearest samples around individual points. For a fixed
$k$, these local relations need not connect the entire point cloud. For
example, when the samples form two well-separated dense clusters, all $k$
nearest neighbors of each sample may lie within the same cluster, leaving the
two clusters disconnected.

The $H_0$ skeleton has a fundamentally different property. By
Proposition~\ref{prop1}, its edge set coincides with that of an MST. An MST
necessarily connects all samples into a single connected structure.
Consequently, every pair of samples in the point cloud is connected by a path
in the $H_0$ skeleton, regardless of whether the samples lie in widely
separated regions. In the two-cluster example above, the MST must contain an
edge crossing the gap between the clusters even when no such relation appears
in a fixed-$k$ nearest-neighbor graph. Thus, local neighborhood relations may
remain confined within disconnected regions of a representation, whereas the
$H_0$ skeleton necessarily connects these regions into a structure spanning
the entire representation.

The following proposition makes this distinction explicit.

\begin{proposition} \textbf{(Global connectivity beyond fixed local neighborhoods).}
Let $k \geq 1$ be fixed, and for every finite point set $Z$ with
$|Z| \geq k+1$, let $G_k(Z)$ denote its symmetrized
$k$-nearest-neighbor graph. Then there exists a finite point set
$X \subset \mathbb{R}$ such that $G_k(X)$ has exactly two connected
components, whereas the $H_0$ skeleton $H_0(X)$ is connected.
Moreover, $H_0(X)$ contains an edge joining the two connected
components of $G_k(X)$ that is not contained in $G_k(X)$.
\label{prop2}
\end{proposition}

\paragraph{Proof.}
Fix $k \geq 1$ and consider
\[
A = \{0,1,\ldots,k\},
\qquad
B = \{3k+2,3k+3,\ldots,4k+2\},
\]
and let $X=A\cup B$.

Each of $A$ and $B$ contains exactly $k+1$ points. Their within-cluster
diameters satisfy
\[
\operatorname{diam}(A)=\operatorname{diam}(B)=k,
\]
whereas the minimum distance between the two clusters is
\[
d(A,B)
=
\min_{a\in A,\,b\in B}|a-b|
=
(3k+2)-k
=
2k+2 > k.
\]
Therefore, for every point in either cluster, its $k$ nearest neighbors
are precisely the other $k$ points in the same cluster. Hence the
subgraphs induced by $A$ and $B$ are both complete, and there is no
edge between them. Thus,
\[
G_k(X)\cong K_{k+1}\sqcup K_{k+1},
\]
where $K_{k+1}$ denotes the complete graph on $k+1$ vertices and
$\sqcup$ denotes the disjoint union. In particular, $G_k(X)$ has
exactly two connected components, namely $A$ and $B$.

Now consider the cut $(A,B)$ of the complete weighted graph on $X$,
with Euclidean distance as the edge weight. The unique minimum-weight
edge crossing this cut is
\[
e^\ast=\{k,\,3k+2\}.
\]
Since $e^\ast$ is the unique minimum-weight edge crossing the cut
$(A,B)$, the cut property of MSTs implies that $e^\ast$ belongs to
every MST of $X$~\citep{kruskal1956shortest}. By
Proposition~\ref{prop1}, the labeled $H_0$ death-edge set coincides
with the edge set of the MST selected under the corresponding
deterministic ordering. Hence $e^\ast \in E_{H_0}^X$.
Since $e^\ast$ joins $A$ and $B$, while $G_k(X)$ contains no edge
between these two components, $e^\ast \notin E(G_k(X))$.
Therefore, the fixed-$k$ neighborhood graph remains disconnected,
whereas the $H_0$ skeleton necessarily contains an edge connecting
its two components and is connected over the entire point set.
\hfill $\square$

This example captures the sense in which $H_0$ skeleton overlap provides
a global counterpart to mKNN. mKNN compares relations defined within
sample-centered local neighborhoods, while the $H_0$ skeleton compares
relations that collectively connect the entire sample set. More generally,
under Kruskal's construction, whether an edge belongs to the $H_0$ skeleton
depends on the connectivity induced by the lower-weight edges processed
before it: an edge is selected precisely when it joins two previously
disconnected components. Thus, the relational support of the $H_0$ skeleton
is determined at the level of the full point cloud rather than independently
around individual samples.

Accordingly, we use mKNN to measure local relational structure and
$H_0$ skeleton overlap to measure global relational structure. Here,
``global'' refers to connectivity across the entire sample set:
individual MST edges need not themselves be geometrically long-range, but
together they form a connected structure spanning all samples.

\section{Additional Relational Structure Results}
\label{app:support}

We provide family-wise analyses complementing the aggregate relational structure
results in Section~\ref{sec:ex_support}.
Figure~\ref{fig:support_convergence} groups vision models by their relative size
within each family and averages the corresponding small, medium, and large models
across vision-model families.
Here, we disaggregate these averages and report the individual variants within
each family.
This analysis tests whether the capacity-dependent relational alignment observed
in the main results is broadly shared across model families rather than being
driven by a particular family.

\subsection{Family-Wise Results}
\label{app:support_family}

\begin{figure}[!ht]
    \centering
    \includegraphics[width=\linewidth]{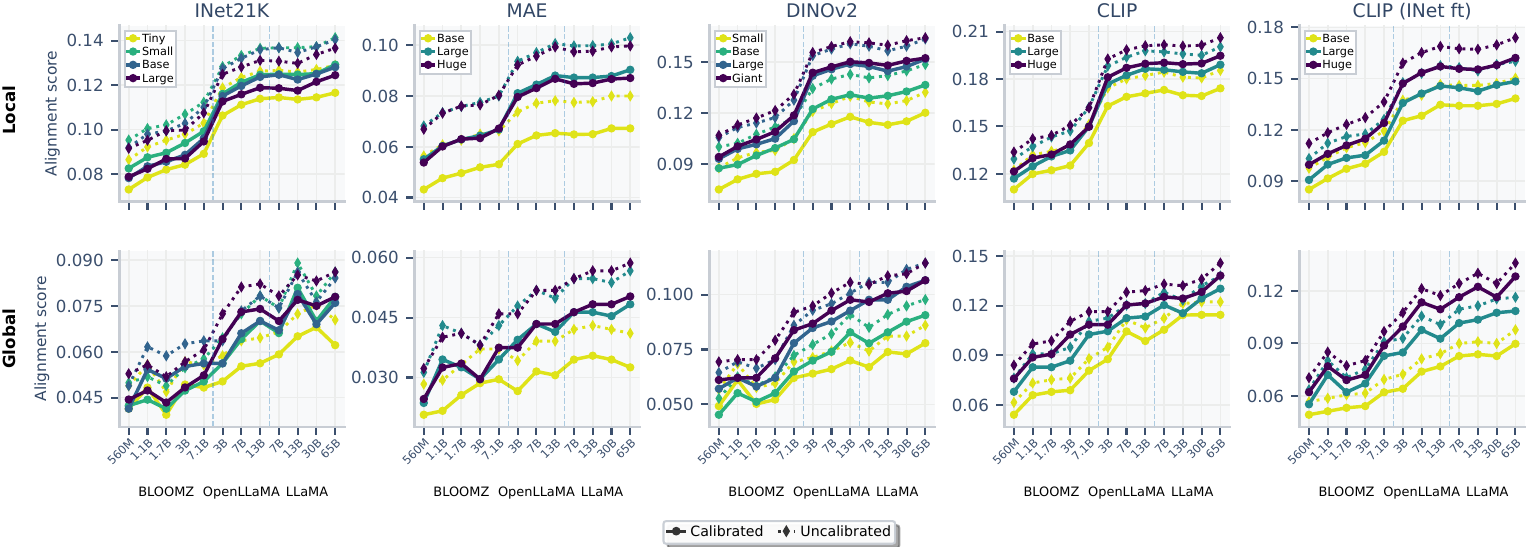}
    \caption{
    \textbf{Family-wise relational structure alignment in vision-language
    representations.}
    Top: local relational structure measured by mKNN.
    Bottom: global spanning structure measured by $H_0$ skeleton overlap.
    Columns correspond to the ImageNet-21K supervised, MAE, DINOv2, CLIP,
    and ImageNet-12K fine-tuned CLIP families, with colors distinguishing
    individual vision-model variants within each family.
    Solid lines with circles show aggregation-aware calibrated alignment,
    and dotted lines with diamonds show uncalibrated alignment.
    Text-model parameter counts are shown along the horizontal axis.
    Each panel uses an independently scaled vertical axis to emphasize the
    within-family capacity-dependent pattern.
    }
    \label{fig:support_family}
\end{figure}

Figure~\ref{fig:support_family} reports local mKNN and global $H_0$ skeleton
overlap separately for the ImageNet-21K supervised, MAE, DINOv2, CLIP, and
ImageNet-12K fine-tuned CLIP families.
Individual vision-model variants are shown within each family, together with
both aggregation-aware calibrated and uncalibrated alignment.

Across families, the local and global measures of relational structure exhibit
broadly similar capacity-dependent patterns.
Alignment generally increases with language-model capacity, and larger
vision-model variants tend to exhibit stronger alignment within each family.
Aggregation-aware calibration reduces the absolute alignment scores but largely
preserves these qualitative relationships.
Importantly, the same capacity-dependent pattern is visible in global $H_0$
spanning structure as in local neighborhood relations.
The aggregate convergence in relational structure reported in the main text is
therefore not attributable to a particular vision-model family, but is broadly
reproduced within the constituent vision-model families.

\section{Additional Distance-Aware Results}
\label{app:value_sensitive}

We provide additional results complementing the aggregate distance-aware
analysis in Section~\ref{sec:ex_value}. We first present the full
capacity-dependent distance-agreement sweep shown compactly in
Figure~\ref{fig:motiv}\textcolor{red}{b}. We then disaggregate the analysis by vision-model
family to examine whether the aggregate pattern arises from averaging across
heterogeneous representation families. Together, these analyses show how the
capacity-dependent signature of convergence changes as distance agreement
becomes increasingly stringent and how consistently this pattern appears
across model families.

\subsection{Full Distance-Agreement Sweep}
\label{app:value_sensitive_sweep}

\begin{figure}[!ht]
    \centering
    \includegraphics[width=\linewidth]{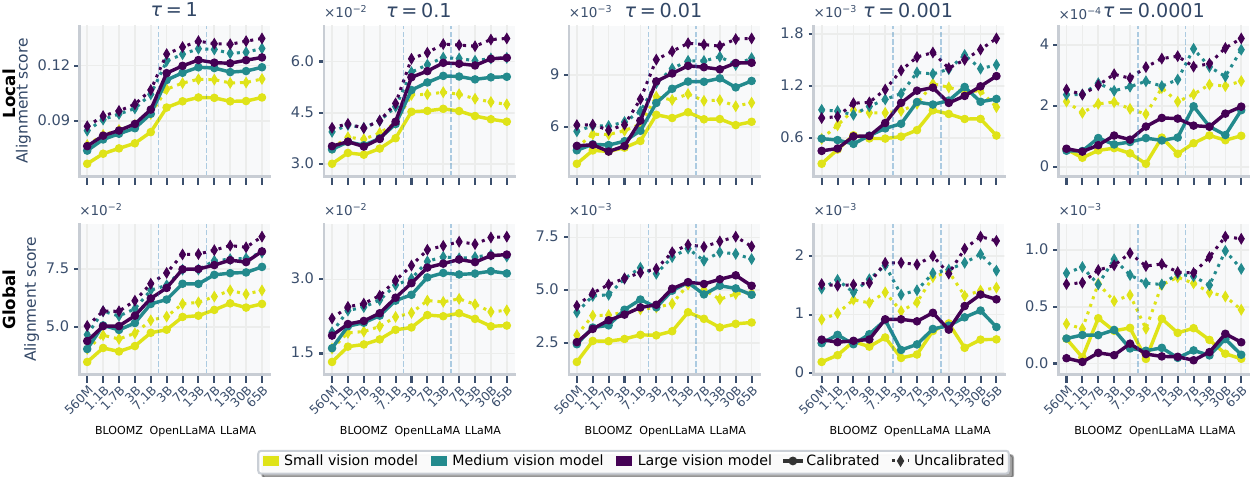}
    \caption{
    \textbf{Full capacity-dependent distance-agreement sweep in
    vision-language representations.}
    Top: local distance-aware mKNN. Bottom: global distance-aware $H_0$
    skeleton overlap. Columns progressively strengthen the
    distance-agreement requirement from $\tau=1$ to $10^{-4}$.
    Small, medium, and large denote relative model sizes within each
    vision-model family, averaged across vision-model families.
    Solid lines with circles show aggregation-aware calibrated alignment,
    and dotted lines with diamonds show uncalibrated alignment.
    Each panel uses an independently scaled vertical axis to visualize the
    within-$\tau$ capacity pattern; the cross-$\tau$ decrease in alignment
    magnitude and statistical significance is quantified in
    Table~\ref{tab:value_sensitive_results}.
    }
    \label{fig:value_sensitive_full_sweep}
\end{figure}

Figure~\ref{fig:value_sensitive_full_sweep} presents an enlarged view of the
full distance-agreement sweep. As in the relational structure analysis, small,
medium, and large denote relative model sizes within each vision-model family,
with corresponding size categories averaged across families. Both
aggregation-aware calibrated and uncalibrated alignment are shown.

Under relatively tolerant distance agreement, the capacity-dependent pattern
observed when comparing relational structure remains clearly visible. At
$\tau=1$ and, to a lesser extent, $\tau=10^{-1}$, alignment generally
increases with language-model capacity, while larger vision models tend to
exhibit stronger alignment for both local distance-aware mKNN and global
distance-aware $H_0$ skeleton overlap. As $\tau$ decreases further, however,
calibrated alignment weakens substantially and the separation associated with
model capacity becomes progressively less stable and less pronounced. Under
the most stringent distance-agreement settings, the calibrated trajectories
are small in magnitude and no longer exhibit the clear capacity-dependent
ordering observed when only relational structure is compared.

Importantly, this transition occurs at both structural scales. The local and
global measures differ in their detailed trajectories, particularly under the
most stringent settings, but neither retains the robust capacity-dependent
pattern observed for relational structure. Thus, the full capacity curves
reinforce the aggregate result in Section~\ref{sec:ex_value}: increasingly
stringent distance agreement weakens the capacity-dependent signature of
representational convergence at both local and global scales.

\begin{figure}[!t]
    \centering

    \begin{subfigure}{\linewidth}
        \centering
        \includegraphics[width=\linewidth]{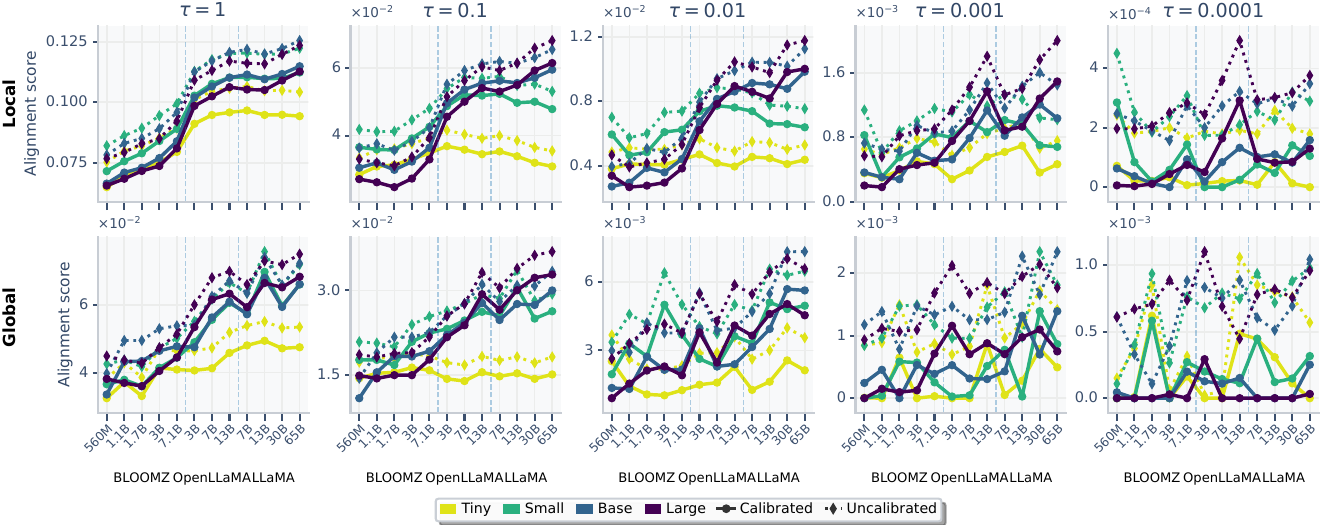}
        \caption{ImageNet-21K supervised models.}
        \label{fig:value_family_inet}
    \end{subfigure}

    \vspace{0.5em}

    \begin{subfigure}{\linewidth}
        \centering
        \includegraphics[width=\linewidth]{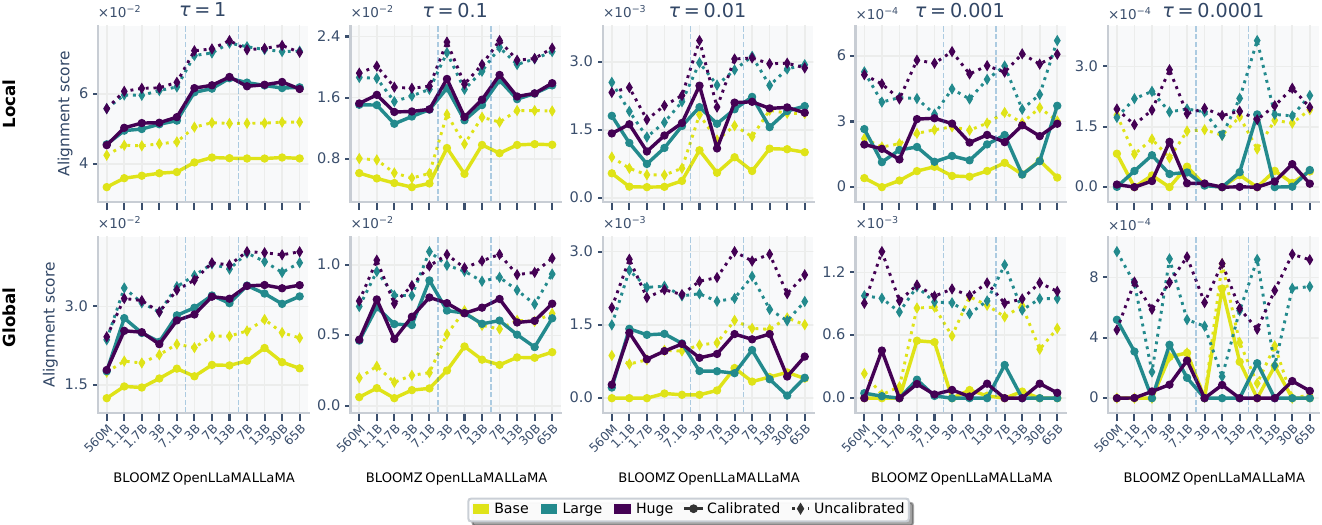}
        \caption{MAE models.}
        \label{fig:value_family_mae}
    \end{subfigure}

    \vspace{0.5em}

    \begin{subfigure}{\linewidth}
        \centering
        \includegraphics[width=\linewidth]{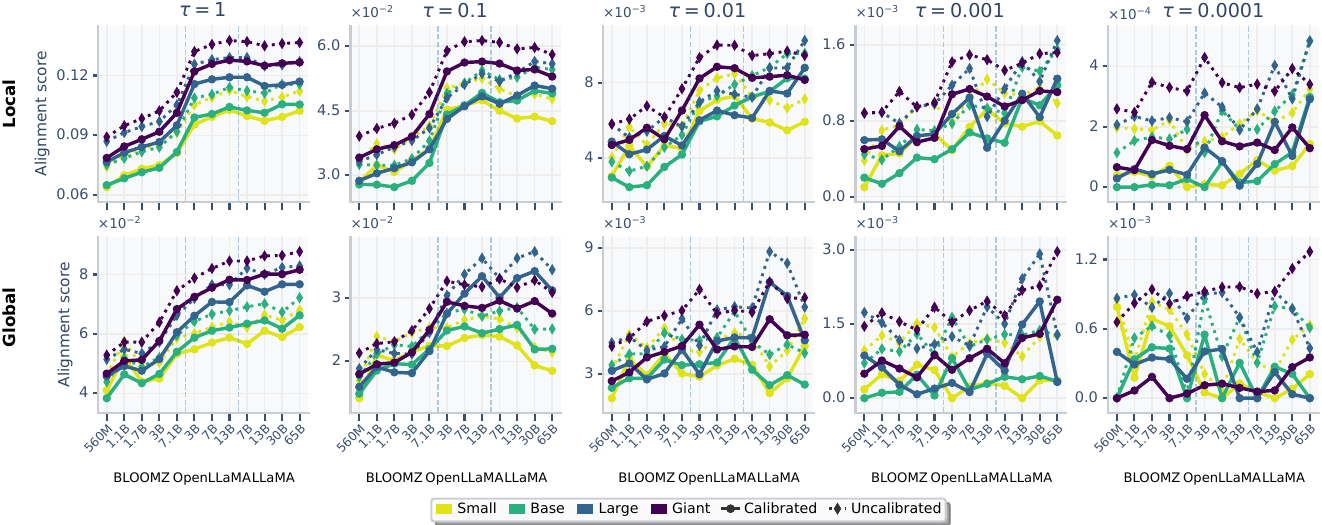}
        \caption{DINOv2 models.}
        \label{fig:value_family_dinov2}
    \end{subfigure}

    \caption{
    \textbf{Family-wise distance-agreement sweeps in vision-language
    representations.}
    Top and bottom rows within each subfigure show local distance-aware
    mKNN and global distance-aware $H_0$ skeleton overlap, respectively.
    Columns vary the distance-agreement requirement from $\tau=1$ to
    $10^{-4}$. Solid lines with circles show aggregation-aware calibrated
    alignment, while dotted lines with diamonds show uncalibrated alignment. Each panel uses an independently scaled vertical axis.
    }
    \label{fig:value_familywise}
\end{figure}

\begin{figure}[!t]
    \ContinuedFloat
    \centering

    \begin{subfigure}{\linewidth}
        \centering
        \includegraphics[width=\linewidth]{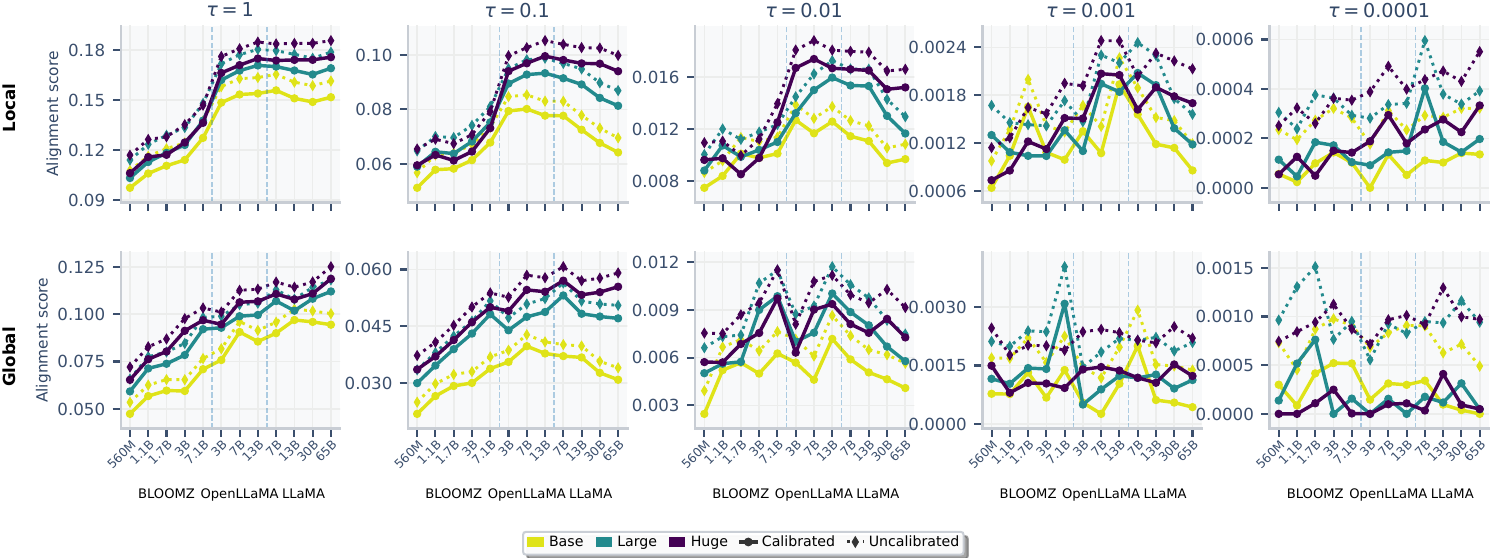}
        \caption{CLIP models.}
        \label{fig:value_family_clip}
    \end{subfigure}

    \vspace{0.5em}

    \begin{subfigure}{\linewidth}
        \centering
        \includegraphics[width=\linewidth]{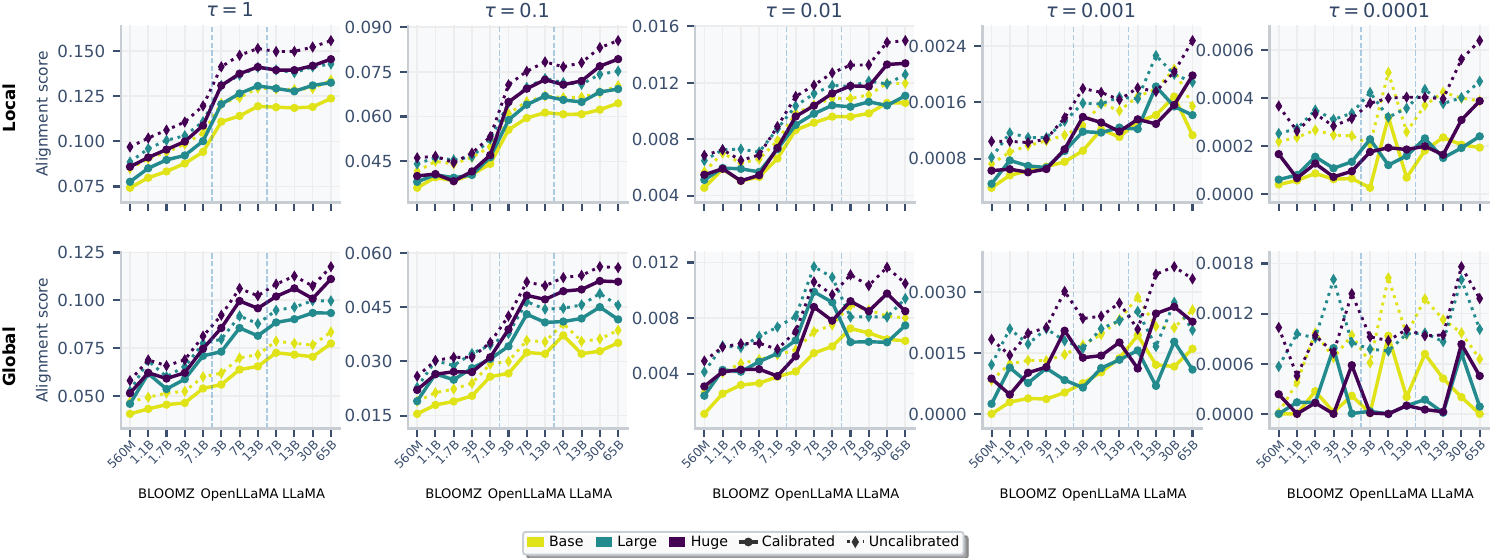}
        \caption{ImageNet-12K fine-tuned CLIP models.}
        \label{fig:value_family_clipft}
    \end{subfigure}

    \caption[]{\textbf{Family-wise distance-agreement sweeps
    (continued).}}
\end{figure}

\begin{table}[!ht]
\centering
\caption{
\textbf{Aggregate distance-agreement results in vision-language representations.}
Relative alignment is the mean aggregation-aware calibrated alignment across
204 model pairs, normalized by the corresponding relational structure mean.
Significant pairs are determined after Benjamini-Hochberg correction.
}
\label{tab:value_sensitive_results}
\begin{tabular}{lcccc}
\toprule
& \multicolumn{2}{c}{Relative calibrated alignment}
& \multicolumn{2}{c}{FDR-significant pairs} \\
\cmidrule(lr){2-3}
\cmidrule(lr){4-5}
Setting & mKNN & $H_0$ & mKNN & $H_0$ \\
\midrule
Support
& 1.0000 & 1.0000
& 204/204 (100.0\%)
& 204/204 (100.0\%) \\

$\tau = 1$
& 0.8560 & 0.8295
& 204/204 (100.0\%)
& 204/204 (100.0\%) \\

$\tau = 10^{-1}$
& 0.3918 & 0.3487
& 204/204 (100.0\%)
& 204/204 (100.0\%) \\

$\tau = 10^{-2}$
& 0.0578 & 0.0547
& 204/204 (100.0\%)
& 201/204 (98.5\%) \\

$\tau = 10^{-3}$
& 0.00681 & 0.00919
& 204/204 (100.0\%)
& 174/204 (85.3\%) \\

$\tau = 10^{-4}$
& 0.000807 & 0.002327
& 183/204 (89.7\%)
& 82/204 (40.2\%) \\
\bottomrule
\end{tabular}
\end{table}

Table~\ref{tab:value_sensitive_results} quantifies the cross-$\tau$ decay
that is not directly comparable from
Figure~\ref{fig:value_sensitive_full_sweep}. Relative calibrated alignment
decreases by more than sixfold between $\tau=10^{-1}$ and $10^{-2}$ at
both structural scales and continues to decline rapidly thereafter.
In contrast, a clear local-global difference emerges primarily in statistical
significance under stringent agreement: at $\tau=10^{-4}$, 89.7\% of local
model pairs remain significant, compared with 40.2\% of global pairs.
Together, these results show that stricter distance agreement produces the
dominant weakening at both scales, while structural scale becomes more
consequential mainly in the stringent regime.

\subsection{Family-Wise Distance-Agreement Sweeps}
\label{app:value_sensitive_family}

We next disaggregate the distance-agreement sweep by vision-model family.
Figure~\ref{fig:value_familywise} shows the complete
$\tau$ sweep separately for ImageNet-21K supervised models, MAE, DINOv2,
CLIP, and ImageNet-12K fine-tuned CLIP. Unlike
Figure~\ref{fig:value_sensitive_full_sweep}, which averages models of
comparable relative size across families, each subfigure retains the individual
vision-model variants within a single family.

The family-wise results broadly reproduce the aggregate transition. At
relatively tolerant values of $\tau$, most families retain a recognizable
capacity-dependent pattern at both local and global scales, with larger
vision-model variants generally exhibiting stronger alignment. As distance
agreement becomes more stringent, calibrated alignment decreases substantially
and the within-family capacity ordering becomes progressively less pronounced
or less stable. The precise transition varies across families: some families
retain a clearer capacity-dependent pattern at intermediate values of $\tau$,
whereas others become irregular earlier. Under the most stringent settings,
however, the calibrated trajectories are weak and substantially less structured
across all families.

These family-specific differences do not alter the main qualitative result.
Both local and global distance-aware alignment weaken as distance agreement
becomes more stringent within the different representation families considered
here. The aggregate weakening in Section~\ref{sec:ex_value} therefore cannot
be explained simply by averaging across heterogeneous vision-model families.
Instead, the loss of robust capacity-dependent alignment under increasingly
stringent distance agreement is broadly reproduced within the constituent
families themselves.

\subsection{Statistical Validation of the Capacity-Dependent Trend}
\label{app:capacity_trend}

The full distance-agreement sweeps in Appendix~\ref{app:value_sensitive_sweep}
show that the capacity-dependent alignment pattern becomes less pronounced
as the agreement requirement becomes more stringent.
Because the absolute magnitude of the distance-aware score is expected to
decrease as $\tau$ decreases, we additionally test whether the association
between model capacity and alignment itself becomes weaker.
We quantify this association using Spearman's rank correlation~\citep{spearman1904proof}, which depends only on the ordering of the
alignment values and is therefore invariant to a uniform rescaling of
scores at a given $\tau$.

\paragraph{Within-family capacity association.}
We evaluate capacity dependence separately along the language- and
vision-model axes.
For language capacity, we hold the vision model fixed and compute
Spearman's rank correlation between model capacity and alignment within
each language-model family (BLOOMZ, OpenLLaMA, and LLaMA).
This yields 51 within-family comparisons from 17 fixed vision models and three
language-model families.
For vision capacity, we hold the language model fixed and compute
Spearman's rank correlation between within-family vision-model size and
alignment separately for ImageNet-21K supervised models, MAE, DINOv2,
CLIP, and ImageNet-12K fine-tuned CLIP.
This yields 60 within-family comparisons from 12 fixed language models and five
vision-model families.
For each structural scale and distance-agreement setting, we summarize
the capacity association by the mean Spearman correlation across valid
within-family comparisons.

Because the same models occur across multiple fixed-model comparisons,
we do not treat these comparisons as independent observations.
Instead, we use an exact permutation test that preserves this model-reuse
structure by permuting capacity labels at the model-family level~\citep{good2005permutation}.
For language capacity, capacity labels are permuted independently within
BLOOMZ, OpenLLaMA, and LLaMA, and each resulting assignment is applied
jointly across all fixed vision models and all values of $\tau$.
This gives
$5!\times3!\times4!=17{,}280$
possible assignments.
For vision capacity, the within-family size labels are permuted jointly
across all fixed language models, giving
$4!\times3!\times4!\times3!\times3!=124{,}416$
possible assignments.
We enumerate all assignments in both analyses.

We use two complementary one-sided statistics.
First, the endpoint statistic measures the decrease in mean capacity
association between the most tolerant and most stringent distance-aware
settings,
\[
T_{\mathrm{end}}
=
\bar{\rho}_{\tau=1}
-
\bar{\rho}_{\tau=10^{-4}}.
\]
Second, we test whether capacity association becomes weaker across the
complete distance-agreement sweep.
Because
$\tau\in\{1,10^{-1},10^{-2},10^{-3},10^{-4}\}$
is evenly spaced on the log scale, we regress the mean Spearman
correlation on the corresponding stringency index
$0,1,2,3,4$
and use the negative slope as the test statistic.
Exact permutation $p$-values are obtained from the complete family-level
label-permutation distribution.
Benjamini-Hochberg correction~\citep{benjamini1995controlling}
is applied across the four language/vision $\times$ local/global
comparisons separately for the endpoint and full-sweep tests.

\begin{table}[!ht]
\centering
\caption{
Statistical validation of the capacity-dependent trend for
aggregation-aware calibrated alignment.
Entries under Support and $\tau$ denote the mean within-family Spearman
correlation between model capacity and alignment.
Support is shown for reference and is not included in the endpoint or
full-sweep tests.
The endpoint decrease is defined as
$\Delta\rho=\bar{\rho}_{\tau=1}-\bar{\rho}_{\tau=10^{-4}}$.
Exact $p$-values are obtained from family-level capacity-label
permutations that preserve model reuse across within-family comparisons.
Benjamini-Hochberg adjusted $q$-values are computed across the four
language/vision $\times$ local/global comparisons separately for the
endpoint and full-sweep tests.
}
\label{tab:capacity_trend_calibrated}
\small

\begin{tabular}{llrrrrrrr}
\toprule
Capacity axis & Scale
& Support
& $\tau=1$
& $10^{-1}$
& $10^{-2}$
& $10^{-3}$
& $10^{-4}$
& $\Delta\rho$ \\
\midrule
Language & Local
& .913 & .735 & .376 & .269 & .351 & .293 & .442 \\
Language & Global
& .770 & .645 & .388 & .353 & .319 & .128 & .517 \\
Vision & Local
& .755 & .775 & .707 & .623 & .478 & .363 & .412 \\
Vision & Global
& .889 & .850 & .722 & .590 & .414 & -.174 & 1.024 \\
\bottomrule
\end{tabular}

\vspace{2mm}

\begin{tabular}{llrrrr}
\toprule
Capacity axis & Scale
& Endpoint $p$
& Endpoint $q$
& Full-sweep $p$
& Full-sweep $q$ \\
\midrule
Language & Local
& .0156 & .0156 & .0369 & .0369 \\
Language & Global
& .0150 & .0156 & .0288 & .0369 \\
Vision & Local
& .00962 & .0156 & .00593 & .0119 \\
Vision & Global
& $1.61{\times}10^{-5}$
& $6.43{\times}10^{-5}$
& $1.61{\times}10^{-5}$
& $6.43{\times}10^{-5}$ \\
\bottomrule
\end{tabular}
\end{table}

\paragraph{Capacity trends after aggregation-aware calibration.}
Table~\ref{tab:capacity_trend_calibrated} shows the within-family
capacity associations for aggregation-aware calibrated alignment.
For the support-only measures, capacity shows a strong positive
association with alignment along both model axes.
The mean Spearman correlations are $0.913$ locally and $0.770$ globally
for language capacity, and $0.755$ and $0.889$, respectively, for vision
capacity.

Within the distance-aware sweep, these positive capacity associations
become substantially weaker as the agreement requirement becomes more
stringent.
For language capacity, the mean local correlation decreases from
$0.735$ at $\tau=1$ to $0.293$ at $\tau=10^{-4}$.
The intermediate trajectory is not strictly decreasing at every value
of $\tau$, but the overall decrease is significant under the endpoint
test
($\Delta\rho=0.442$, exact $p=0.0156$,
BH-adjusted $q=0.0156$)
and is also significant across the full sweep
($p=0.0369$, $q=0.0369$).
For global alignment, the mean language-capacity correlation decreases
from $0.645$ to $0.128$.
The endpoint decrease is again significant
($\Delta\rho=0.517$, $p=0.0150$, $q=0.0156$),
as is the weakening across the complete sweep
($p=0.0288$, $q=0.0369$).

The same pattern appears along the vision-capacity axis.
Mean local correlation decreases from $0.775$ at $\tau=1$ to
$0.363$ at $\tau=10^{-4}$
($\Delta\rho=0.412$, exact $p=0.00962$, $q=0.0156$),
with a significant decrease across the full sweep
($p=0.00593$, $q=0.0119$).
For global alignment, the mean correlation decreases from $0.850$
to $-0.174$, giving the largest endpoint decrease among the four
comparisons
($\Delta\rho=1.024$, $p=1.61\times10^{-5}$,
$q=6.43\times10^{-5}$).
The full-sweep test is also significant
($p=1.61\times10^{-5}$, $q=6.43\times10^{-5}$).

For the vision-capacity/global comparison, the negative mean correlation
at the strictest setting should not be interpreted as evidence for a
general inverse scaling relationship.
Rather, it indicates that the strong positive capacity ordering observed
under the support-only measure and tolerant distance agreement is no
longer preserved under this stringent agreement requirement.

These results separate the weakening of the capacity-dependent pattern
from the mechanically expected decrease in alignment magnitude.
Spearman correlation depends only on the ordering of alignment values.
Therefore, if decreasing $\tau$ merely multiplied all alignment scores
by a smaller common factor, the capacity association would remain
unchanged.
Instead, the rank association between model capacity and alignment
itself becomes substantially weaker at both structural scales and along
both model-capacity axes.

\section{Robustness to Normalization and Similarity-Based Comparison}
\label{app:construction_robustness}

\subsection{Sensitivity to Distance Normalization}
\label{app:distance_normalization}

Our main distance-aware analysis normalizes each pairwise distance matrix by
the 90th percentile of its positive distances before evaluating distance
agreement. To test whether the observed pattern depends on this particular
normalization choice, we repeat the full analysis using the 50th, 75th, and
95th percentiles as alternative reference scales. Specifically, we replace
the normalization factor $Q_{0.9}(\{D_{ij}:D_{ij}>0\})$ with
$Q_q(\{D_{ij}:D_{ij}>0\})$ for
$q\in\{0.50,0.75,0.95\}$, while keeping the local and global structural
supports, the $\tau$ sweep, aggregation-aware calibration, permutation testing,
and multiple-testing correction unchanged.

\begin{figure}[!ht]
    \centering
    \includegraphics[width=\linewidth]{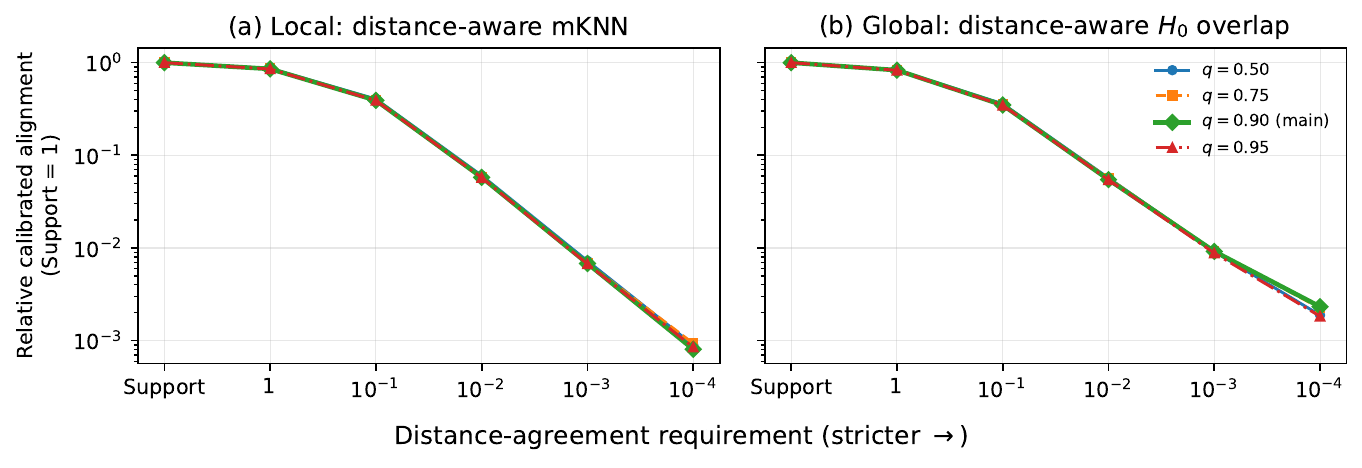}
    \caption{
    \textbf{Sensitivity to distance normalization.}
    Relative aggregation-aware calibrated alignment under alternative
    distance-normalization quantiles. The 90th percentile ($q=0.90$) is the main
    setting used throughout the paper, while $q\in\{0.50,0.75,0.95\}$ provides
    alternative normalization scales. Across both
    \textbf{(a)} local distance-aware mKNN and
    \textbf{(b)} global distance-aware $H_0$ skeleton overlap,
    the alignment trajectories remain nearly unchanged across normalization
    choices as the distance-agreement requirement becomes more stringent.
    }
    \label{fig:distance_normalization}
\end{figure}

Figure~\ref{fig:distance_normalization} shows that the resulting alignment
trajectories are highly consistent across normalization quantiles. For local
distance-aware mKNN, the alternative normalizations retain approximately
85-87\% of the corresponding relational structure baseline at $\tau=1$ and
only about 6\% at $\tau=10^{-2}$. Global distance-aware $H_0$ skeleton overlap
exhibits nearly the same transition, retaining approximately 83-84\% at
$\tau=1$ and about 5-6\% at $\tau=10^{-2}$. Under still more stringent
agreement, alignment approaches zero for every normalization choice. The
trajectories closely track the main 90th-percentile setting throughout the
full $\tau$ sweep.

\begin{table}[!ht]
\centering
\caption{
\textbf{FDR-significant model pairs under alternative distance-normalization
quantiles.}
Values denote the number of significant pairs out of 204 after
Benjamini-Hochberg correction.
}
\label{tab:distance_normalization_fdr}
\small
\begin{tabular}{c c ccccc}
\toprule
$q$ & Scale
& $1$ & $10^{-1}$ & $10^{-2}$ & $10^{-3}$ & $10^{-4}$ \\
\midrule
0.50
& Local  & 204 & 204 & 204 & 204 & 182 \\
& Global & 204 & 204 & 201 & 186 & 84  \\
\midrule
0.75
& Local  & 204 & 204 & 204 & 202 & 176 \\
& Global & 204 & 204 & 202 & 184 & 109 \\
\midrule
0.90 (main)
& Local  & 204 & 204 & 204 & 204 & 183 \\
& Global & 204 & 204 & 201 & 174 & 82  \\
\midrule
0.95
& Local  & 204 & 204 & 204 & 201 & 181 \\
& Global & 204 & 204 & 201 & 181 & 58  \\
\bottomrule
\end{tabular}
\end{table}

Table~\ref{tab:distance_normalization_fdr} shows that the significance results
preserve the same qualitative pattern. All 204 model pairs remain significant
at both structural scales through $\tau=10^{-1}$ under every normalization
choice. At $\tau=10^{-2}$, all local pairs remain significant, while
201-202 global pairs remain significant. Differences become more pronounced
only under stringent agreement: at $\tau=10^{-4}$, 176-182 local pairs and
58-109 global pairs remain significant across the alternative normalization
quantiles. Although the exact significance counts vary in this regime, global
alignment consistently becomes less prevalent than local alignment.

Together, these results show that the weakening under increasingly stringent
distance agreement, as well as the greater global fragility in the stringent
regime, is robust to the choice of distance-normalization quantile.

\subsection{Similarity-Aware Alignment}
\label{app:similarity_values}

Our main analysis evaluates agreement using distances assigned to the shared
relations. We next examine whether the observed weakening persists under an
analogous similarity-based construction by replacing distance agreement with
cosine-similarity agreement while keeping the underlying local and global
structural supports fixed. For a shared relation $(i,j)$, let
\[
s_X(i,j)
=
\frac{x_i^\top x_j}{\|x_i\|\,\|x_j\|},
\qquad
s_Y(i,j)
=
\frac{y_i^\top y_j}{\|y_i\|\,\|y_j\|}.
\]
We replace the distance-agreement weight in
\eqref{eq:distance_weight} with
\begin{equation}
w_{ij,\mathrm{sim}}^{(\tau)}
=
\exp\left(
-\frac{
\left|s_X(i,j)-s_Y(i,j)\right|
}{\tau}
\right),
\label{eq:similarity_weight}
\end{equation}
where smaller $\tau$ requires closer agreement in cosine similarity.
The weight is applied only to relations shared by the two representations,
using the same local kNN and global $H_0$ skeleton supports as in the main
analysis. As $\tau\rightarrow\infty$, the weights approach one and the
corresponding relational structure measures are recovered. We otherwise retain
the same $\tau$ sweep, aggregation-aware calibration, permutation testing, and
multiple-testing correction.

\begin{figure}[!ht]
    \centering
    \includegraphics[width=\textwidth]{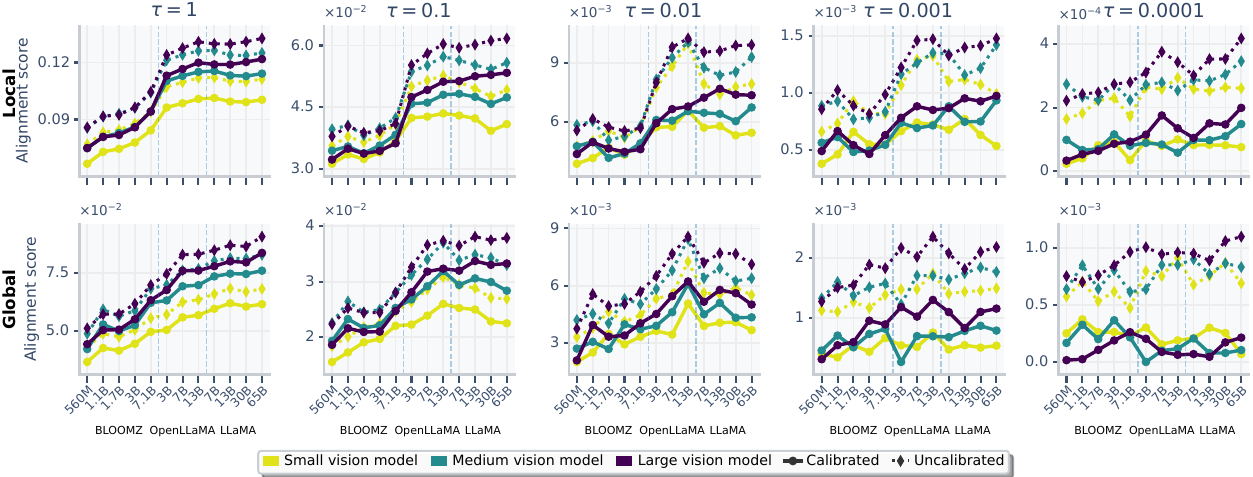}
    \caption{
    \textbf{Full similarity-aware alignment sweep.}
    Top and bottom rows show local similarity-aware mKNN and global
    similarity-aware $H_0$ skeleton overlap, respectively. Columns strengthen
    the cosine-similarity agreement requirement from $\tau=1$ to $10^{-4}$.
    Solid-circle and dotted-diamond lines denote calibrated and uncalibrated
    alignment. As agreement becomes more stringent, calibrated alignment weakens
    and the capacity-dependent pattern becomes less pronounced at both scales.
    Each panel uses an independently scaled vertical axis.
    }
    \label{fig:similarity_sweep}
\end{figure}

Figure~\ref{fig:similarity_sweep} shows the full similarity-based sweep.
Under relatively tolerant similarity agreement, both local and global measures
retain a clear capacity-dependent structure, with larger vision models generally
exhibiting stronger alignment. As the agreement requirement becomes more
stringent, calibrated alignment decreases substantially and the capacity-dependent
pattern becomes progressively less pronounced at both structural scales. Each
panel uses an independently scaled vertical axis to expose the within-$\tau$
capacity pattern; the corresponding decrease in alignment magnitude is
summarized in Table~\ref{tab:similarity_results}.

\begin{table}[!ht]
\centering
\caption{
\textbf{Numerical results for similarity-aware alignment.}
Relative alignment is the mean aggregation-aware calibrated alignment across
204 model pairs, normalized by the corresponding relational structure mean.
Significant pairs are determined after Benjamini-Hochberg correction.
}
\label{tab:similarity_results}
\begin{tabular}{lcccc}
\toprule
& \multicolumn{2}{c}{Relative calibrated alignment}
& \multicolumn{2}{c}{FDR-significant pairs} \\
\cmidrule(lr){2-3}
\cmidrule(lr){4-5}
Setting & mKNN & $H_0$ & mKNN & $H_0$ \\
\midrule
Support
& 1.0000 & 1.0000
& 204/204 (100.0\%)
& 204/204 (100.0\%) \\

$\tau = 1$
& 0.8437 & 0.8489
& 204/204 (100.0\%)
& 204/204 (100.0\%) \\

$\tau = 10^{-1}$
& 0.3554 & 0.3522
& 204/204 (100.0\%)
& 204/204 (100.0\%) \\

$\tau = 10^{-2}$
& 0.0473 & 0.0557
& 204/204 (100.0\%)
& 204/204 (100.0\%) \\

$\tau = 10^{-3}$
& 0.00571 & 0.00957
& 203/204 (99.5\%)
& 196/204 (96.1\%) \\

$\tau = 10^{-4}$
& 0.000763 & 0.002628
& 190/204 (93.1\%)
& 123/204 (60.3\%) \\
\bottomrule
\end{tabular}
\end{table}

The similarity-aware construction closely reproduces the main transition in
Table~\ref{tab:value_sensitive_results}. At $\tau=1$, local and global
alignment retain 84.4\% and 84.9\% of their relational structure baselines,
compared with 85.6\% and 83.0\% under the distance-aware construction. At
$\tau=10^{-2}$, the corresponding similarity-aware values fall to 4.73\% and
5.57\%, closely paralleling the 5.78\% and 5.47\% retained under distance
agreement. Under still more stringent requirements, alignment approaches zero
at both structural scales. Thus, the qualitative weakening persists when
agreement on the shared relations is evaluated using cosine similarity rather
than distance.

The exact significance trajectories show somewhat greater dependence on the
choice of relation value. Similarity-aware alignment remains significant for
more model pairs in the stringent regime, particularly at the global scale:
196 and 123 global pairs remain significant at $\tau=10^{-3}$ and
$10^{-4}$, compared with 174 and 82 under the distance-aware construction.
Nevertheless, the same qualitative pattern remains: alignment weakens at both
structural scales as agreement becomes more stringent, while significant
global alignment becomes less prevalent than local alignment in the stringent
regime. These results show that the observed weakening is not specific to the
particular distance-agreement construction used in the main analysis.

\section{Robustness under a Locally Estimated Riemannian Metric}
\label{app:riemannian}

Our main analysis shows robust convergence in relational structure, whereas
requiring increasingly stringent agreement in the associated Euclidean
distances progressively weakens alignment. This raises an alternative
explanation: the observed weakening may depend specifically on the ambient
Euclidean metric. We therefore repeat the distance-aware analysis using a
locally estimated Riemannian metric.

Crucially, we keep the local kNN and global MST structural supports fixed and
replace only the distances used to evaluate the corresponding relations.
Reconstructing the supports under the Riemannian metric would change both
relational structure and metric information simultaneously, reintroducing the
confound that our framework is designed to avoid. The experiment therefore
tests whether the same structure-geometry pattern persists when distance
agreement is evaluated under an alternative, locally adaptive metric.

\subsection{Riemannian Distance Construction}
\label{app:riemannian_method}

We construct a locally covariance-adapted Riemannian metric to test whether
the observed distance-agreement pattern depends on the ambient Euclidean
metric~\citep{singer2008non,berry2016local}. The construction is applied
independently to each representation layer. We first $\ell_2$-normalize
the representation vectors and, for each sample $x_i$, identify its
$k_R$ nearest neighbors in the ambient space. In our experiments, we use
$k_R=30$.
The resulting symmetrized $k_R$-NN graph was connected for all
representation layers considered in our experiments.

Because the normalized representations lie on the unit sphere, we map
each neighbor $x_j$ to the tangent space at $x_i$ using the spherical
logarithm map
\[
v_{ij}
=
\log_{x_i}(x_j)
=
\frac{\theta_{ij}}{\sin\theta_{ij}}
\left(
x_j-\cos\theta_{ij}\,x_i
\right),
\qquad
\theta_{ij}
=
\arccos(x_i^\top x_j).
\]

Let $V_i$ denote the matrix whose rows are the tangent vectors
$\{v_{ij}:j\in\mathcal{N}_{k_R}(i)\}$. We estimate the local covariance
structure as
\[
C_i=\frac{1}{k_R}V_i^\top V_i
\]
and retain its leading $d_R$ eigenvectors
$q_{i1},\ldots,q_{id_R}$ with corresponding eigenvalues
$\lambda_{i1},\ldots,\lambda_{id_R}$. These eigenvectors define the
estimated local tangent subspace
\[
\widehat{\mathcal T}_i
=
\operatorname{span}\{q_{i1},\ldots,q_{id_R}\}.
\]
Let $P_i$ denote the orthogonal projection onto
$\widehat{\mathcal T}_i$. On this subspace, we define the regularized
local quadratic form
\[
g_i(u,u)
=
\sum_{r=1}^{d_R}
\frac{\langle u,q_{ir}\rangle^2}
{\lambda_{ir}+\gamma\bar{\lambda}_i},
\qquad
\bar{\lambda}_i
=
\frac{1}{d_R}\sum_{r=1}^{d_R}\lambda_{ir},
\qquad
u\in\widehat{\mathcal T}_i,
\]
where $\gamma=10^{-3}$ controls covariance regularization and
$d_R=10$.

For an undirected edge $(i,j)$, we project the endpoint-local tangent
displacements onto their respective estimated tangent subspaces and
symmetrize the resulting quadratic forms. We define the Riemannian edge
length as
\[
\ell_R(i,j)
=
\left[
\frac{1}{2}
\left(
g_i(P_i v_{ij},P_i v_{ij})
+
g_j(P_j v_{ji},P_j v_{ji})
\right)
\right]^{1/2}.
\]
These lengths define a symmetric weighted graph on the union of the
ambient $k_R$-nearest-neighbor edges.

We use this locally estimated metric differently for the local and global
distance-aware measures while keeping their structural supports fixed.
For distance-aware mKNN, the original $k=10$ neighbor identities are retained
and only the distances assigned to these relations are replaced by
$\ell_R(i,j)$. The edge lengths are normalized by the 90th percentile of the
positive Riemannian graph-edge lengths before applying the same log-distance
agreement weighting used in the main analysis.

For distance-aware $H_0$ skeleton overlap, the original MST edge identities
remain those obtained from the ambient Euclidean metric. To obtain Riemannian
distances for these globally spanning relations, we compute all-pairs
shortest-path distances on the symmetric Riemannian-weighted $k_R$-NN graph,
\[
D_R(i,j)
=
\min_{\pi:i\rightsquigarrow j}
\sum_{(u,v)\in\pi}\ell_R(u,v),
\]
and normalize the positive entries of $D_R$ by their 90th percentile.
The resulting values $D_R(i,j)$ are then evaluated only on the fixed MST
support. Thus, the Riemannian experiment changes the metric used to evaluate
distance agreement while preserving the local and global relational
structures being compared.

\subsection{Full Riemannian Results}
\label{app:riemannian_results}

\begin{figure}[!ht]
    \centering
    \includegraphics[width=\linewidth]{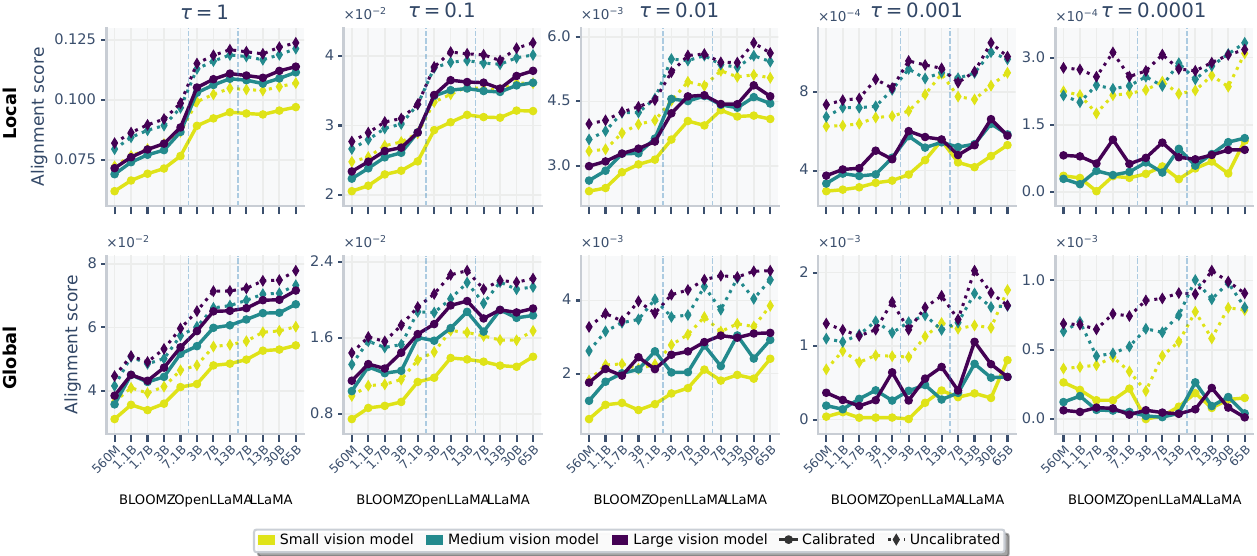}
    \caption{
    \textbf{Full capacity-dependent alignment patterns under a locally estimated
    Riemannian metric.}
    Top: local distance-aware mKNN. Bottom: global distance-aware $H_0$ skeleton
    overlap. Columns progressively strengthen the distance-agreement requirement
    from $\tau=1$ to $10^{-4}$. Solid lines show aggregation-aware calibrated
    alignment, and dotted lines show uncalibrated alignment. As distance agreement
    becomes more stringent, calibrated alignment weakens and the capacity-dependent
    pattern becomes less pronounced at both structural scales. Each panel uses an
    independently scaled vertical axis to visualize the within-$\tau$ capacity
    pattern; the cross-$\tau$ decrease in alignment magnitude is summarized in
    Table~\ref{tab:riemannian_results} and
    Figure~\ref{fig:robustness_replication}\textcolor{red}{a}.
    }
    \label{fig:riemannian_sweep}
\end{figure}

Figure~\ref{fig:riemannian_sweep} shows the full capacity-dependent
alignment patterns under the locally estimated Riemannian metric.
Under relatively tolerant distance agreement, both local and global measures
retain the capacity-dependent pattern observed when only relational structure
is compared. As the distance-agreement requirement becomes more stringent,
however, calibrated alignment weakens and the capacity-dependent pattern
becomes progressively less pronounced at both structural scales.
This qualitative behavior closely parallels the pattern obtained under the
ambient Euclidean metric.

Table~\ref{tab:riemannian_results} quantifies this decay. At
$\tau=1$, the Riemannian variants retain $79.0\%$ and $72.4\%$ of the
corresponding relational structure baselines for the local and global measures,
respectively. These fractions decrease to $25.8\%$ and $20.4\%$ at
$\tau=10^{-1}$ and to only $3.24\%$ and $2.97\%$ at
$\tau=10^{-2}$. Under still stricter requirements, the retained alignment
falls below $0.5\%$ at $\tau=10^{-3}$ and approaches zero at
$\tau=10^{-4}$. Thus, the pronounced decline is not specific to the ambient
Euclidean metric.

\begin{table}[!ht]
\centering
\caption{
\textbf{Distance-aware results under a locally estimated Riemannian metric.}
Relative alignment is the mean aggregation-aware calibrated alignment
across 204 model pairs normalized by the corresponding relational structure
mean. Euclidean relative alignment is included for reference. Significant
pairs are reported as counts and percentages after Benjamini-Hochberg
correction.
}
\label{tab:riemannian_results}
\small
\begin{tabular}{c cc cc cc}
\toprule
& \multicolumn{2}{c}{Local}
& \multicolumn{2}{c}{Global}
& \multicolumn{2}{c}{Riemannian significant pairs} \\
\cmidrule(lr){2-3}
\cmidrule(lr){4-5}
\cmidrule(lr){6-7}
$\tau$
& Euclidean & Riemannian
& Euclidean & Riemannian
& Local & Global \\
\midrule
Support
& 1.000 & 1.000
& 1.000 & 1.000
& 204/204 (100\%) & 204/204 (100\%) \\
$1$
& 0.856 & 0.790
& 0.830 & 0.724
& 204/204 (100\%) & 204/204 (100\%) \\
$10^{-1}$
& 0.392 & 0.258
& 0.349 & 0.204
& 204/204 (100\%) & 204/204 (100\%) \\
$10^{-2}$
& 0.0578 & 0.0324
& 0.0547 & 0.0297
& 204/204 (100\%) & 203/204 (99.5\%) \\
$10^{-3}$
& 0.00681 & 0.00394
& 0.00919 & 0.00488
& 204/204 (100\%) & 160/204 (78.4\%) \\
$10^{-4}$
& 0.000807 & 0.000525
& 0.002327 & 0.001399
& 175/204 (85.8\%) & 13/204 (6.4\%) \\
\bottomrule
\end{tabular}
\end{table}

Permutation significance exhibits the same overall progression. All
204 model pairs remain significant at both structural scales through
$\tau=10^{-1}$. At $\tau=10^{-2}$, all local pairs and 203 of 204
global pairs remain significant. Under stricter agreement, the global
measure becomes more fragile: at $\tau=10^{-3}$, 204 local pairs but
160 global pairs remain significant, and at $\tau=10^{-4}$ the counts
decrease to 175 and 13, respectively. Thus, increasingly stringent distance
agreement weakens alignment at both scales, with an additional scale-dependent
difference emerging in the stringent regime.

Taken together, these results show that replacing the ambient Euclidean metric
with a locally covariance-adapted Riemannian metric does not restore the robust
capacity-dependent convergence observed when only relational structure is
compared. The same overall structure-geometry pattern therefore persists
under this alternative metric.

\section{Additional Video-Text Results}
\label{app:video_text}

We provide additional results for the video-text extension described in
Section~\ref{sec:video_text}. We first examine relational structure alignment
across individual visual-model families and then report the full
distance-agreement sweep. Together, these analyses show that the qualitative
structure-geometry pattern observed in the main vision-language setting
extends to video-text representations: alignment in relational structure
remains robust at both structural scales, whereas increasingly stringent
distance agreement progressively weakens alignment and the capacity-dependent
pattern.

For the capacity-dependent plots, we show the 12 language models belonging to
the BLOOMZ, OpenLLaMA, and LLaMA families. The singleton Gemma-2-9B-IT model is
omitted from these plots because it does not define a within-family capacity
trajectory, but it remains included in the aggregate statistics over all
143 video-text model pairs.

\subsection{Family-Wise Relational Structure Results}
\label{app:video_text_support}

\begin{figure}[!ht]
    \centering
    \includegraphics[width=\linewidth]{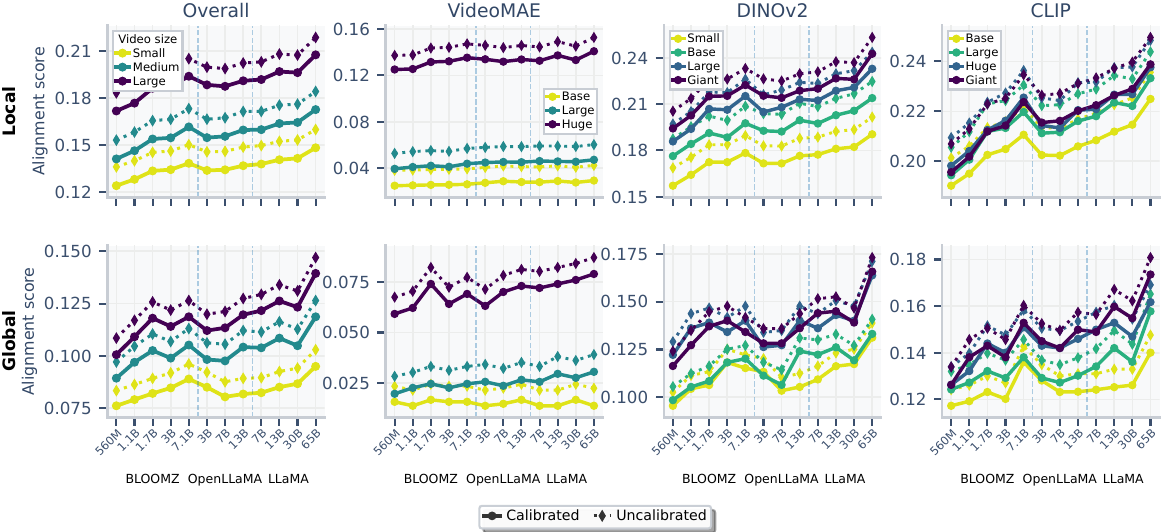}
    \caption{
    \textbf{Family-wise relational structure alignment in video-text representations.}
    Top: local relational structure measured by mKNN. Bottom: global spanning
    structure measured by $H_0$ skeleton overlap. The overall panels summarize
    small, medium, and large relative size categories across VideoMAE, DINOv2, and CLIP, while the remaining panels show individual variants
    within each family. Solid lines with circles show aggregation-aware
    calibrated alignment, and dotted lines with diamonds show uncalibrated
    alignment. Text-model parameter counts are shown along the horizontal axis.
    Each panel uses an independently scaled vertical axis to emphasize the
    within-family capacity-dependent pattern.
    }
    \label{fig:video_text_support_full}
\end{figure}

Figure~\ref{fig:video_text_support_full} expands the relational structure
analysis by visual-model family. The overall panels summarize relative small,
medium, and large model variants across VideoMAE, DINOv2, and CLIP, while the
remaining panels show the individual variants within each family. We report
both aggregation-aware calibrated and uncalibrated alignment.

Across the three visual-model families, local mKNN and global $H_0$ skeleton
overlap exhibit broadly similar capacity-dependent patterns. Calibration
reduces the absolute alignment scores but preserves the qualitative trends:
alignment generally increases with language-model capacity, and larger
visual-model variants tend to exhibit stronger relational alignment. Although
the absolute score ranges differ across VideoMAE, DINOv2, and CLIP, the
corresponding local and global patterns remain qualitatively consistent.
Thus, the robust relational structure alignment observed in the aggregate
analysis is not driven by a single visual-model family.

\subsection{Full Distance-Agreement Sweep}
\label{app:video_text_distance}

\begin{figure}[!ht]
    \centering
    \includegraphics[width=\linewidth]{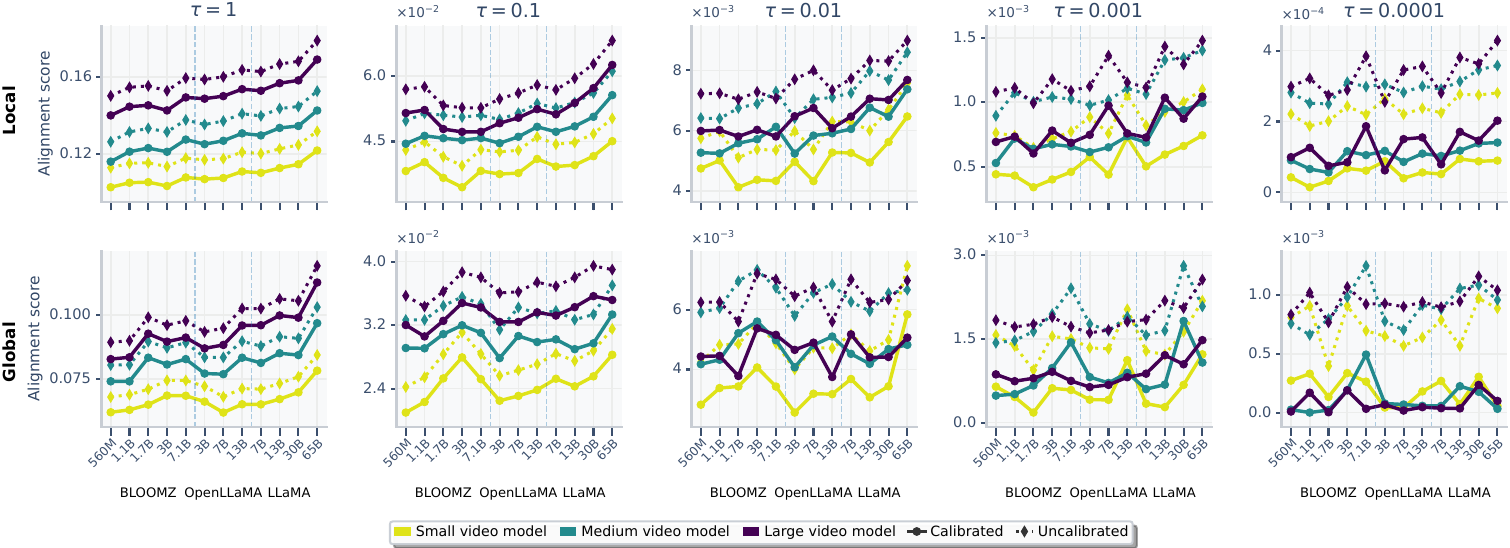}
    \caption{
    \textbf{Full distance-aware alignment sweep in video-text representations.}
    Top: local distance-aware mKNN. Bottom: global distance-aware $H_0$ skeleton
    overlap. Columns progressively strengthen the distance-agreement requirement
    from $\tau=1$ to $10^{-4}$. Colors denote relative visual-model size,
    averaged across VideoMAE, DINOv2, and CLIP. Solid lines with circles show
    aggregation-aware calibrated alignment, and dotted lines with diamonds show
    uncalibrated alignment. Each panel uses an independently scaled vertical axis
    to visualize the within-$\tau$ capacity pattern; the cross-$\tau$ decrease
    in alignment magnitude is summarized in
    Table~\ref{tab:video_text_full_results} and
    Figure~\ref{fig:robustness_replication}\textcolor{red}{(b,c)}.
    }
    \label{fig:video_text_tau_full}
\end{figure}

Figure~\ref{fig:video_text_tau_full} shows the full capacity-dependent
video-text results as the distance-agreement requirement is progressively
strengthened. At relatively tolerant values of $\tau$, both local
distance-aware mKNN and global distance-aware $H_0$ skeleton overlap retain
the capacity-dependent pattern observed when only relational structure is
compared. As $\tau$ decreases, however, calibrated alignment decreases sharply
and the separation associated with model capacity becomes progressively less
pronounced at both structural scales.

Importantly, this transition occurs in parallel for the local and global
measures. The full capacity curves therefore support the aggregate result in
Section~\ref{sec:video_text}: increasingly stringent distance agreement weakens
alignment at both structural scales rather than selectively eliminating global
alignment. Differences between the two scales become more apparent only under
stringent distance agreement, particularly in the prevalence of model pairs
that remain statistically significant.

\begin{table}[!ht]
\centering
\caption{
\textbf{Full numerical results for video-text representations.}
Relative alignment is the mean aggregation-aware calibrated alignment across
143 video-text model pairs, normalized by the corresponding relational structure
mean. Significant pairs are reported as counts and percentages after
Benjamini-Hochberg correction.
}
\label{tab:video_text_full_results}
\begin{tabular}{lcccc}
\toprule
& \multicolumn{2}{c}{Relative calibrated alignment}
& \multicolumn{2}{c}{FDR-significant pairs} \\
\cmidrule(lr){2-3}
\cmidrule(lr){4-5}
Setting & mKNN & $H_0$ & mKNN & $H_0$ \\
\midrule
Support
& 1.0000
& 1.0000
& 143/143 (100.0\%)
& 143/143 (100.0\%) \\

$\tau = 1$
& 0.8109
& 0.7936
& 143/143 (100.0\%)
& 143/143 (100.0\%) \\

$\tau = 10^{-1}$
& 0.2930
& 0.2914
& 143/143 (100.0\%)
& 143/143 (100.0\%) \\

$\tau = 10^{-2}$
& 0.0365
& 0.0422
& 143/143 (100.0\%)
& 142/143 (99.3\%) \\

$\tau = 10^{-3}$
& 0.00439
& 0.00741
& 143/143 (100.0\%)
& 131/143 (91.6\%) \\

$\tau = 10^{-4}$
& 0.000596
& 0.001224
& 123/143 (86.0\%)
& 67/143 (46.9\%) \\
\bottomrule
\end{tabular}
\end{table}

Table~\ref{tab:video_text_full_results} quantifies the corresponding decay.
At $\tau=1$, local and global alignment retain 81.1\% and 79.4\% of their
respective relational structure baselines. These fractions decrease almost
identically to 29.3\% and 29.1\% at $\tau=10^{-1}$, and further to 3.65\%
and 4.22\% at $\tau=10^{-2}$. Thus, the dominant reduction in alignment
remains closely matched across structural scales.

A scale-dependent difference emerges more clearly in statistical significance.
All 143 local model pairs remain significant through $\tau=10^{-3}$, whereas
the number of significant global pairs decreases from 142/143 at
$\tau=10^{-2}$ to 131/143 at $\tau=10^{-3}$. At the most stringent setting,
$\tau=10^{-4}$, 123/143 (86.0\%) local pairs remain significant, compared
with 67/143 (46.9\%) global pairs. These results reproduce the same
structure-geometry pattern observed in the vision-language analysis:
alignment becomes substantially less robust as increasingly stringent distance
agreement is required at both scales, while structural scale contributes an
additional difference mainly in the stringent regime.

\end{document}